# Robust High-Resolution Multi-Organ Diffusion MRI Using Synthetic-Data-Tuned Prompt Learning

Chen Qian[1,#], Haoyu Zhang[2,#], Junnan Ma[3], Liuhong Zhu[4,1], Qingrui Cai[1], Yu Wang[5], Ruibo Song[5], Lv Li[5], Lin Mei[5], Xianwang Jiang[5], Qin Xu[5], Boyu Jiang[6], Ran Tao[6], Chunmiao Chen[7], Shufang Chen[7], Dongyun Liang[4], Qiu Guo[8], Jianzhong Lin[9], Taishan Kang[9], Mengtian Lu[10], Liyuan Fu[11], Ruibin Huang[12], Huijuan Wan[13], Xu Huang[14], Jianhua Wang[14], Di Guo[15], Hai Zhong[3], Jianjun Zhou[4] and Xiaobo Qu[1,*]

**Abstract:** Clinical adoption of multi-shot diffusion-weighted magnetic resonance imaging (multi-shot DWI) for body-wide tumor diagnostics is limited by severe motion-induced phase artifacts from respiration, peristalsis, and so on, compounded by multi-organ, multi-slice, multi-direction and multi-b-value complexities. Here, we introduce a reconstruction framework, LoSP-Prompt, that overcomes these challenges through physics-informed modeling and synthetic-data-driven prompt learning. We model inter-shot phase variations as a high-order Locally Smooth Phase (LoSP), integrated into a low-rank Hankel matrix reconstruction. Crucially, the algorithm's rank parameter is automatically set via prompt learning trained exclusively on synthetic abdominal DWI data emulating physiological motion. Validated across 10,000+ clinical images (43 subjects, 4 scanner models, 5 centers), LoSP-Prompt: 1) Achieved twice the spatial resolution of clinical single-shot DWI, enhancing liver lesion conspicuity; 2) Generalized to 7 diverse anatomical regions (liver, kidney, sacroiliac, pelvis, knee, spinal cord, brain) with a single model; 3) Outperformed state-of-the-art methods in image quality, artifact suppression, and noise reduction (11 radiologists' evaluations on a 5-point scale, $p<0.05$), achieving 4-5 point (excellent) on kidney DWI, 4 points (good to excellent) on liver, sacroiliac and spinal cord DWI, and 3-4 points (good) on knee and tumor brain. The approach eliminates navigator signals and realistic data supervision, providing an interpretable, robust solution for high-resolution multi-organ multi-shot DWI. Its scanner-agnostic performance signifies transformative potential for precision oncology.

**Teaser:** Prompt learning; Synthetic data learning; High-resolution diffusion weighted imaging; Multi-organ.

Diffusion weighted imaging (DWI) in magnetic resonance imaging (MRI) can detect the *in vivo* water molecule movements non-invasively[1], which has been widely employed in clinical diagnosis of tumors[2-4] in brain and abdomen. Compared with the clinically commonly used single-shot echo planar imaging DWI sequence, multi-shot interleaved echo planar imaging (ms-iEPI) sequence greatly improves DWI with higher resolution, better signal-to-noise ratio (SNR), and lower geometric distortion[5-9], bringing great diagnostic values.

However, the ms-iEPI DWI is very sensitive to the inter-shot motion during the data acquisition of each shot (**Fig. 1(a-d)**). Even slight movement on the millimeter scale will cause the significant extra inter-shot phase (motion-induced phase in **Fig. 1(e, f)**) due to the amplification by strong diffusion gradients[10]. The motion-induced phase will disturb the phase encoding, resulting in frequency shift of k-space data and severe motion artifacts on DWI images (**Fig. 1(g)**)[11,12].

The inter-shot motion affects the amplitude and wrapping degree of the motion-induced phase[10] (Appendix **Note 1**).

Previous works show that, the approximate rigid-body translation or rotation motions of the brain result in motion-induced phase composed of smooth functions[13,14] (**Fig. 1(e, f)**). The smooth phase in the image domain is then formulated as the low-rank property of the k-space (Fourier transform of the image), leading to many state-of-the-art (SOTA) multi-shot DWI reconstruction methods, such as low-rank optimization methods (ALOHA[15], MUSSELS[13,16], LORAKS[14,17], PAIR[18], DONATE[19]) and low-rank deep learning method (MoDL-MUSSELS[20]). All these methods can successfully remove image artifacts in brain imaging (**Fig. 1(h)**), greatly promoting applications of multi-shot high-resolution DWI.

For the abdominal tumor diagnosis, such as liver and kidney, ms-iEPI DWI has not been applied well (**Fig. 1(I)**). A main reason is abdomen organs suffer from non-rigid movements and elastic deformations because of varying degrees of physiological movement[21-23], e.g., heartbeat, breathing movement, and intestinal peristalsis (Appendix **Note 1**). These movements bring organ-specific and high-order motion-induced phases (**Fig. 1(i,** j)), which do not conform to the smooth phase prior assumption made in multi-shot DWI brain imaging. For example, in abdominal imaging (**Fig. 1(i-l)**), locally smooth inter-shot phase is presented in the abdomen organs, making the low-rank assumption violated (**Fig. 1(n)**), resulting in serious residual motion artifacts in image reconstruction (**Fig. 1(l)**).

[1]Department of Electronic Science, Fujian Provincial Key Laboratory of Plasma and Magnetic Resonance, National Institute for Data Science in Health and Medicine, Xiamen University, China. [2]Pen-Tung Sah Institute of Micro-Nano Science and Technology, Xiamen University [3]Department of Radiology, The Second Hospital of Shandong University, China. [4]Department of Radiology, Zhongshan Hospital, Fudan University (Xiamen Branch), China. [5]Neusoft Medical Systems, China. [6]United Imaging Healthcare, China. [7]Department of Radiology, Lishui Central Hospital, China. [8]Department of Radiology, Xiang'an Hospital of Xiamen University, School of Medicine, Xiamen University, China. [9]Department of Radiology, Zhongshan Hospital Affiliated to Xiamen University, China. [10]Department of Radiology, Xianning Central Hospital, The First Affiliated Hospital of Hubei University of Science and Technology, China. [11]Department of Radiology, 900th Hospital of PLA Joint Logistic Support Force, Dongfang Hospital of Xiamen University, School of Medicine, Xiamen University [12]Department of Radiology, The First Affiliated Hospital of the Medical College of Shantou University, China. [13]Department of Neurology and Department of Neuroscience, The First Affiliated Hospital of Xiamen University, School of Medicine, Xiamen University, Xiamen, China. [14]Department of Radiology, The First Affiliated Hospital of Xiamen University, China. [15]School of Computer and Information Engineering, Xiamen University of Technology, China.
\# is equal contribution. *Correspondence should be addressed to Xiaobo Qu (quxiaobo@xmu.edu.cn).

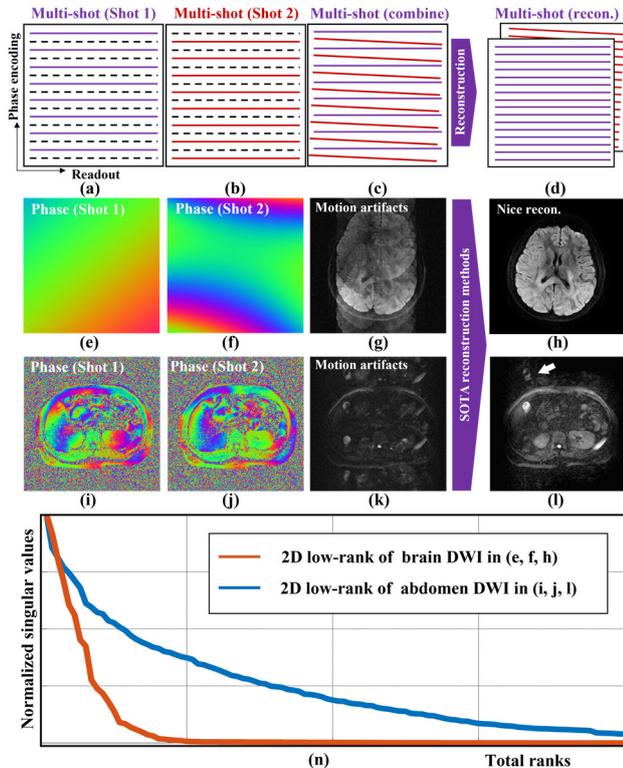

**Fig. 1 | Multi-shot interleaved echo planar imaging DWI reconstruction in brain and abdomen. (a, b)** are the sampled k-space of the 1$^{st}$ and 2$^{nd}$ shot, respectively. **(c)** is the combined k-space without reconstruction. **(d)** is the reconstructed k-space of 2 shots. **(e, i)** and **(f, j)** are the phases of the 1$^{st}$ and 2$^{nd}$ shot, respectively. **(g, k)** are unreconstructed DWI images. **(h, l)** are reconstructed DWI images with a 2D low-rank method (PAIR). **(n)** is the singular value attenuation of 2D low-rank matrix of brain and abdomen. Note: The solid and dotted lines represent sampled and unsampled k-space data, respectively.

Thus, existing methods have limited adaptation for high-resolution multi-organ abdominal DWI reconstructions since they hardly consider these complex phase characteristics of moving organs outside the brain.

In this work, we model complex inter-shot phase as the high-order Locally Smooth Phase (LoSP) and propose an organ-specific LoSP reconstruction model with Prompt learning (LoSP-Prompt) for multi-organ high-resolution DWI. This framework consists of two parts. The first part is a 1D low-rank optimization method LoSP, which decomposes the 2D DWI image reconstruction problem into a multiple 1D signal recovery[24,25] along both readout and phase encoding directions. The second part is a prompt learning with a Prompt-Net (modified ResNet18[26]), that learns the critical parameter, saved ranks, of 1D signal recovery from synthesized data, and then automatically apply it to realistic multi-organ DWI reconstruction, thus improving LoSP robustness.

A toy example of the LoSP-Prompt is demonstrated in **Fig. 2**. The 1D signal decomposition decouples the high-order locally smooth phase of specific organ ($\Omega$ in **Fig. 2(a)**) and the low-order locally smooth phase of other organs ($\bar{\Omega}$ in **Fig. 2(a)**). This decoupling preserves the good low-rankness of 1D signals, effectively isolating the rank increase induced by high-order phase (**Fig. 2(c)**). By contrast, traditional 2D low-rank methods inevitably suffer compromised overall low-rankness due to high-order phase (**Fig. 2(b)**), thus making it impossible to reliably distinguish image structures from artifacts (**Fig. 2(h-j)**). Our baseline LoSP, as a 1D decoupling approach, achieves better performance (**Fig. 2(k-m)**) than the 2D low-rank method (**Fig. 2(h-j)**).

Our 1D decoupling approach introduces new complexity: Each decoupled 1D signal recovery requires a distinct number of saved ranks, $r$, for optimal reconstruction (**Fig. 2(d)**). This parameter essentially determines the number of principal components retained in the low-rank approximation of each 1D signal. A higher $r$ allows capturing more signal details and potential high-frequency information, but risking of more noise or residual artifacts (arrow ① in **Fig. 2(k)**). Conversely, a lower $r$ enforces stronger denoising and artifact suppression but may remove weak signals (arrow ② in **Fig. 2(k)**). Setting $r$ is more challenging if high-order phase existed in multi-organ DWI (arrow ③ in **Fig. 2(m)**). Thus, choosing an appropriate $r$ is crucial and must be adapted to the specific characteristics (e.g., noise level, phase complexity) of each 1D signal. Manually configuring these signal-specific parameters across diverse anatomical regions (even multi-organ, multi-slice, multi-direction and multi-b-value) would be prohibitively labor-intensive and clinically impractical.

Prompt learning provides new perspective for mining and utilizing the information from synthetic data[25,27]. Compared to directly using the synthetic data as labels for supervised training[28], a Prompt-Net transfers auxiliary reconstruction information (saved ranks in (**Fig. 2(d)**)) from the synthetic data for reconstructions. This Prompt-Net provides an automatic way to predict these signal-adaptive save ranks, allowing that 1D signals from different regions being subject to differentiated low-rank constraints (blue line in **Fig. 2(d)**), thus greatly improve abdomen DWI reconstruction (**Fig. 2(n-p)**) than the baseline LoSP.

In the following, comprehensive experiments will show that LoSP-Prompt surpasses SOTA methods on three aspects: 1) robust and high-consistent liver DWI and apparent diffusion coefficient (ADC) with twice the resolution of clinical practice (acquisition matrix size is 256×256) (**Fig. 3-5**); 2) nice clinical adaptability to liver lesions (**Fig. 6**) and brain lesions (**Fig. 7**); 3) better artifacts removal and noise suppression in generalized reconstruction of multi-organ DWI (**Fig. 7**). More results (Appendix **Note 5-7**) further reveal the robust reconstruction by LoSP-Prompt for both fully-sampling (Appendix **Fig. S5-14**) and 2 under-sampling (Appendix **Fig. S15-16**).

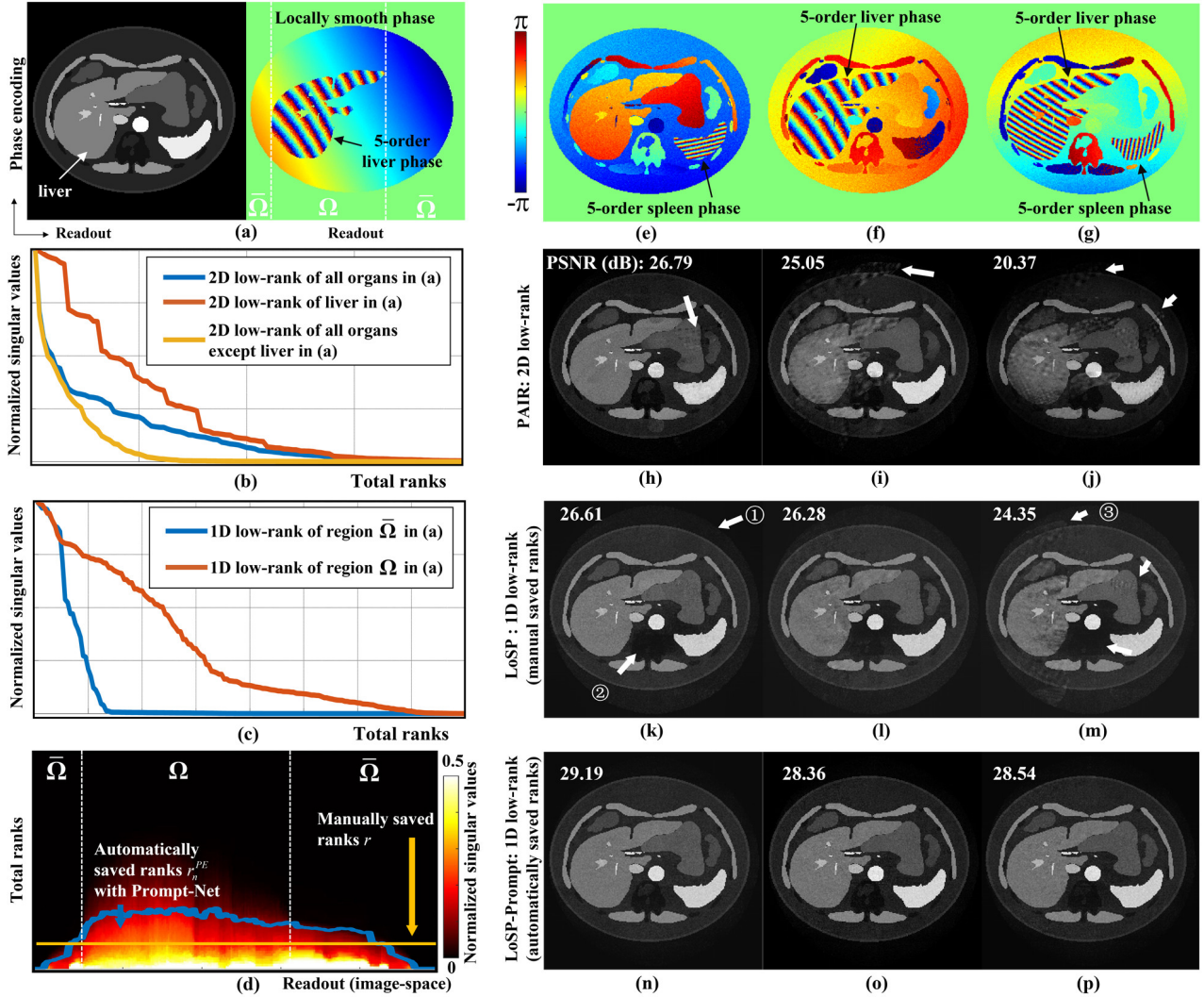

**Fig. 2 | Robustness of LoSP-Prompt to organ-specific locally-smooth phase data. (a)** The abdomen magnitude and locally smooth phase consisted of liver with 5-order liver phase and other organs with 1-order phases. **(b)** are the singular value attenuation curves of 2D low-rank of different organs in **(a)**. **(c)** are the singular value attenuation curves of 1D low-rank of 1D signals from regions $\Omega$ and $\bar{\Omega}$, respectively. **(d)** are the singular value attenuation curves of 1D low-rank of all 1D signals in **(a)**. **(e-g)** are 5-order phases with random polynomial coefficients for **(e)** spleen, **(f)** liver, and **(g)** both, respectively, and other organs (kidney, pancreas, fat, and so on) have 1-order phases. **(h-j)**, **(k-m)**, and **(n-p)** are reconstructed DWI images by PAIR (2D low-rank), the proposed LoSP (1D low-rank with manually saved ranks), and LoSP-Prompt (1D low-rank with automatically saved ranks), respectively. Note: $\Omega$ and $\bar{\Omega}$ in **(a)** are the readout regions including and excluding the liver, respectively; PSNRs are marked at the top of **(h-j, k-m, n-p)**; Only the 1D low rank along the readout is shown in **(d)** for simplicity.

## RESULTS

Here, we will demonstrate the robustness of LoSP-Prompt through comparison with SOTA methods. Three low-rank methods (MUSSELS[13,16], S-LORAKS[14,17], LLR[29]) are set with optimized reconstruction parameters (Appendix **Table. VI**) to balance artifacts removal and noise suppression for each subject. The critical parameter of the proposed method, the save ranks, is manually set in the baseline LoSP for each subject and automatically provided in LoSP-Prompt with Prompt-Net for all subjects. Two home-made *in vivo* ms-iEPI DWI datasets (Appendix **Note 2** for scan parameters) are used, providing a total of more than 10,000 DWI images, from 4 scanner models in 5 centers. **DATASET I (Patient DWI)** contains high-resolution DWI of 3 patients with liver tumors and 18 patients with brain metastases. **DATASET II (Healthy volunteer DWI)** contains high-resolution DWI of 6 body parts, including liver, kidney, sacroiliac, pelvis, knee and spinal cord. Each part contains DWI data from at least 3 healthy volunteers.

## Multi-b-value and multi-slice high-resolution abdomen DWI reconstruction and ADC quantification

Multi-b-value and multi-slice raise challenges of multi-shot DWI image reconstruction. The former introduces different levels of signal-to-noise-ratio (SNR) since the signal intensity reduces exponentially with b-value while the latter may introduce different levels of motions, e.g. stronger motions of liver parts that are closer to hearts.

In multi-b-value reconstructions (**Fig. 3**), the performance of most compared methods vary greatly at different b-values (50 and 800 s/mm$^2$). Under the low b-value (50 s/mm$^2$), all methods provide comparable results. Under the high b-value (800 s/mm$^2$), S-LORAKS leads to obvious noise or motion artifacts residuals (arrow ② in **Fig. 3(b)**), suggesting a loose regularity constraint, while MUSSELS and LLR have signal loss of some liver tissues (arrows ① in **Fig. 3(a, c)**), indicating a tight regularity constraint.

In multi-slice reconstructions (**Fig. 4**), inappropriate regularization constraints of compared methods are more obvious. LLR, S-LORAKS, and MUSSELS achieve nice reconstructions in the 22$^{nd}$ slice, but have artifacts residual (**Fig. 4(b)**) or signal loss (**Fig. 4(a, c)**) in the 8$^{th}$ and 15$^{th}$ slices. These comparisons imply that, compared methods with a set of optimized reconstruction parameters are difficult to meet the reconstruction requirements of abdomen DWI reconstructions (Appendix **Note 5** for all slices).

In both multi-b-value and multi-slice DWI reconstructions, the proposed baseline, LoSP, achieves better robustness than LLR, S-LORAKS, and MUSSELS. However, LoSP still introduces signal loss of liver tissue (**Fig. 3(d) and Fig. 4(d)**). With prompt learning, LoSP-Prompt maintains much better robust ability of noise suppression, motion artifacts removal, and image structure protection across multi-b-value and multi-slice DWIs (**Fig. 3(e)** and **Fig. 4(e)**).

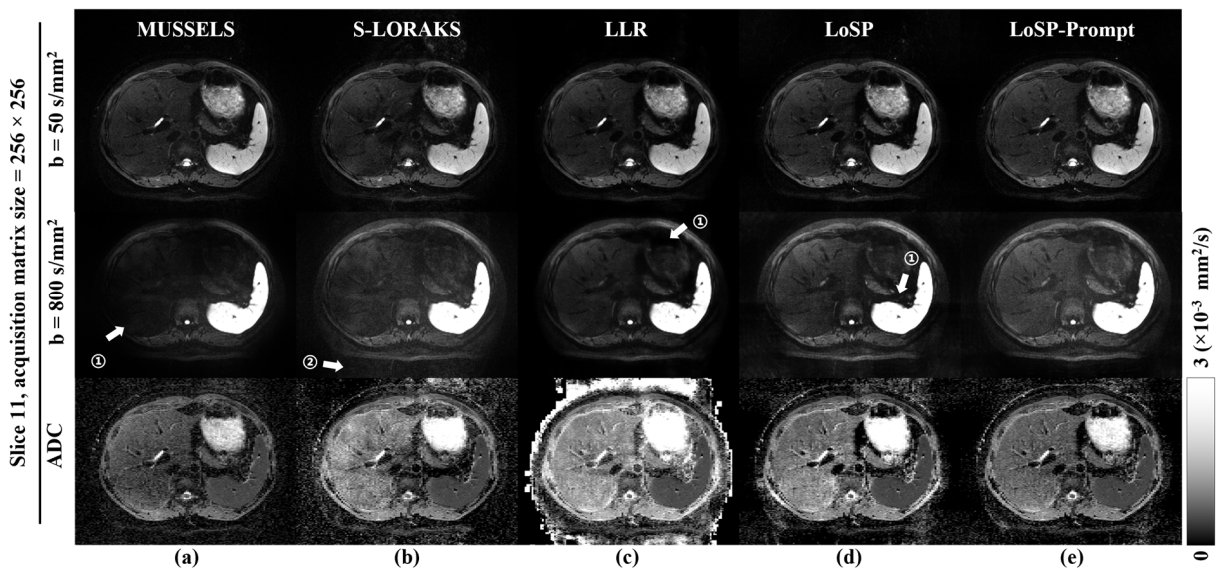

**Fig. 3 | Robustness of LoSP-Prompt to multi-b-value DWI data. (a-e)** are reconstructed DWI images and ADC maps by MUSSELS, S-LORAKS, LLR, LoSP, and LoSP-Prompt, respectively. Note: The data is acquired with United Imaging 5T Jupiter scanner, subject ID is **HS#13**.

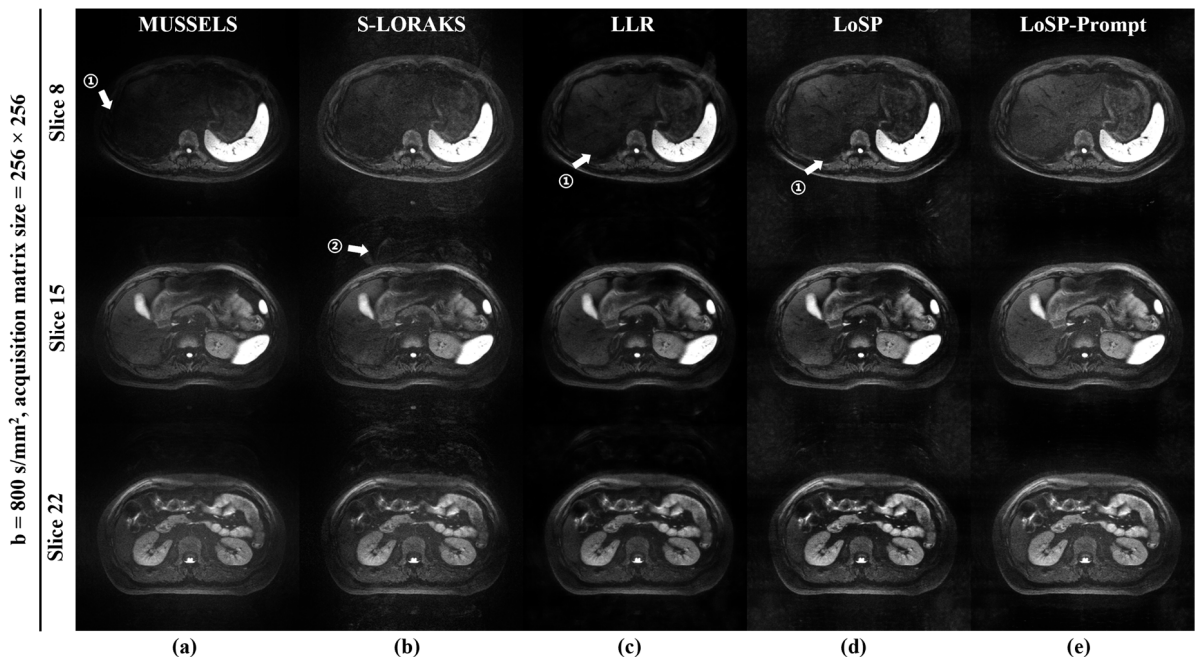

**Fig. 4 | Robustness of LoSP-Prompt to multi-slice DWI data. (a-e)** are reconstructed DWI images and ADC maps by MUSSELS, S-LORAKS, LLR, LoSP, and LoSP-Prompt, respectively. Note: The data is acquired with United Imaging 5T Jupiter scanner, subject ID is **HS#13**.

Apparent diffusion coefficient (ADC), a quantitative biomarker derived from DWI, that can be used to assess cellularity, predict tumor aggressiveness, and monitor treatment response [30,31], is analyzed in two sets of regions of interest (ROIs) in **Fig. 5**. In the first set (**Fig. 5(b)**), the 8 circular ROIs are randomly placed within slice 11. In the second set (**Fig. 5(c-g)**), the 8 circular ROIs are selected from 8 liver segments (**Fig. 5(a)**) in slice 5, 9, 11, 13, and 16. The ADC (1.26±0.14 mm²/s) of normal liver tissue is adopted as the reference [32].

In the intra-slice ADC quantification (**Fig. 5(h)**), MUSSELS and LLR exhibit a tendency to underestimate and overestimate the ADC values, respectively. S-LORAKS, LoSP, and LoSP-Prompt yield most ADC values falling in the reference range of normal liver tissue. The LoSP-Prompt achieves best quantification consistency and the highest overall agreement with reference range.

In the inter-slice ADC quantification (**Fig. 5(i)**), MUSSELS produces ADC values mostly below the reference range, while LLR shows a tendency to overestimate ADC values. Comparing with MUSSELS and LLR, both S-LORAKS and LoSP lead to more consistent ADC with the reference range. But S-LORAKS and LoSP encounters the problem of large fluctuations of ADC in different liver segments. Notably, LoSP-Prompt improved quantification consistency than the baseline LoSP, and achieved the highest overall agreement with reference liver ADC values among all methods.

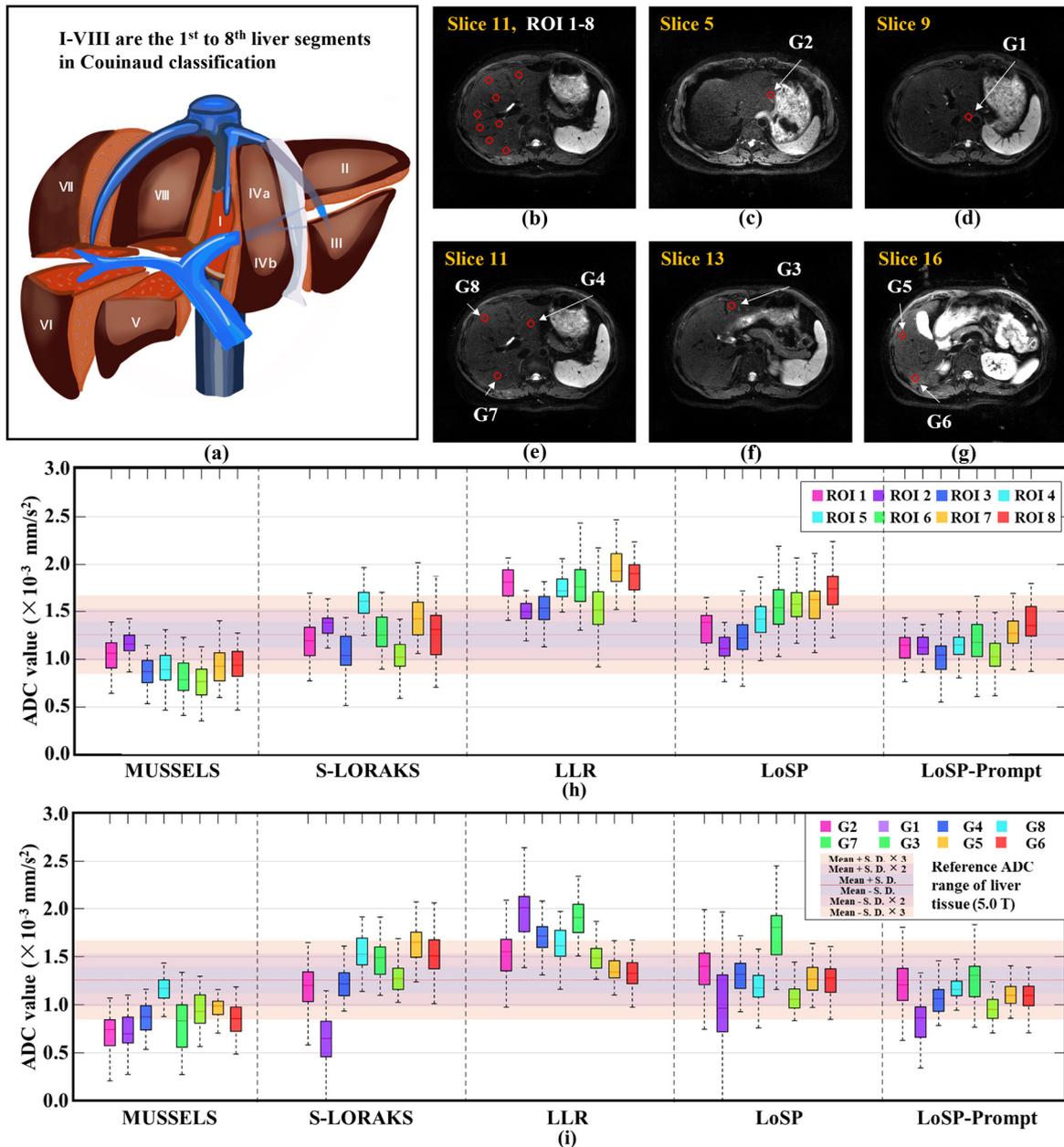

**Fig. 5 | Consistency of LoSP-Prompt to ADC quantifications. (a)** the eight functionally independent segments of liver (Couinaud classification). **(b)** a group of ROIs selected in one slice. **(c-g)** a group of ROIs selected in eight liver segments of five slices. **(h)** the distribution of ADC in eight ROIs of **(b)**. **(i)** the distribution of ADC in selected eight ROIs of **(c-g)**. Note: S. D. is short for standard deviation. For each ROI, the diameter has 9 pixels. The data is acquired with United Imaging 5T Jupiter scanner, subject ID is **HS#13**.

Thus, LoSP-Prompt achieves better robustness to multi-value and multi-slice DWI and better quantitative consistency of ADC values than compared methods.

**Clinical adaptability to patient data with liver lesions**

Here, we validate the clinical adaptability of LoSP-Prompt on patient data with liver lesions (**Fig. 6**).

The lesion in the left lateral lobe of the liver has uneven slightly long T2 signals (yellow arrows in the 1$^{st}$ row of **Fig. 6(b)**), with a cross-section of about 2.1×1.9 cm. In the dynamic contrast-enhanced MRI (2$^{nd}$ and 3$^{rd}$ rows of **Fig. 6(a)**), the lesion is enhanced obviously in the arterial phase, and the contrast agent withdrew in the venous phase, showing relatively low signals. These imaging features indicate that the lesion is highly suspected of hepatocellular carcinoma. After a puncture biopsy, pathology confirms it to be well-differentiated hepatocellular carcinoma (Appendix **Note 11**).

The multi-shot DWI provides better SNR than single-shot DWI under the same acquisition resolution, which brings better lesion detectability. In the single-shot DWI (arrow ① in **Fig. 6(c)**), the lesion is not obvious, thus analyzing its signal strength is hard. In the multi-shot DWI (arrow ① in **Fig. 6(d, e)**), the lesion is easier to distinguish from liver tissue, and its higher signal than around tissue indicating its malignant potential, matching the radiographic features of hepatocellular carcinoma. Moreover, compared with single-shot DWI, multi-shot DWI can significantly reduce the liver deformation (arrows ③ in **Fig. 6(c, e)**). The lower distortion is helpful for analyzing the morphology and properties of lesion.

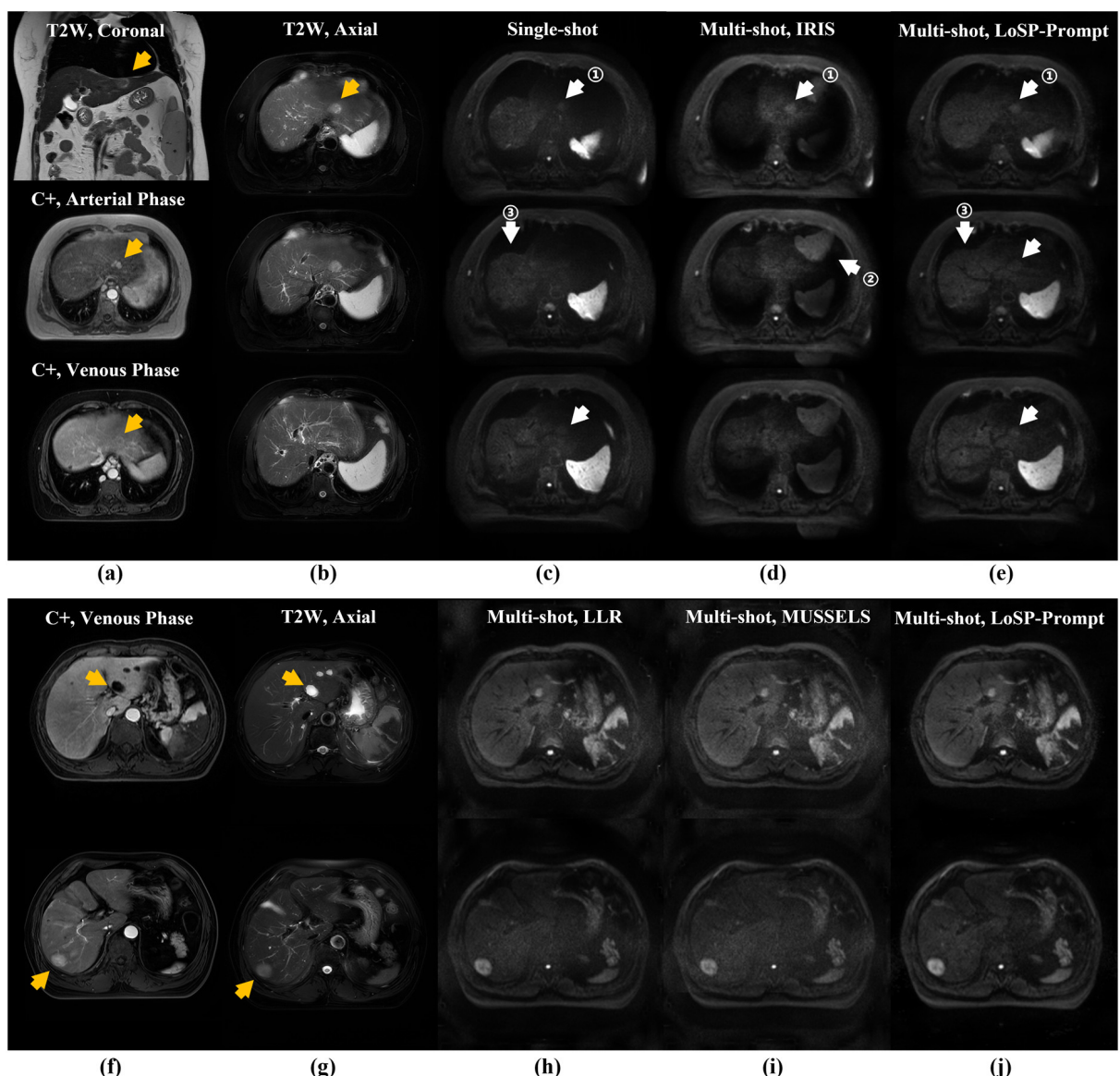

**Fig. 6 | Clinical adaptability to liver patient DWI reconstructions.** (a)-(b) are references, including T2-weighted, contrast (C+) enhanced T1-weighted of arterial and venous phases. (c) is single-shot DWI. (d)-(e) are multi-shot DWI reconstructed by navigator method IRIS, and navigator-free method LoSP-Prompt. (f)-(g) are the references, including T2-weighted, enhanced T1-weighted of venous phase. (h)-(j) are the multi-shot DWI reconstructed by LLR, MUSSELS, and LoSP-Prompt, respectively. Note: The multi-shot DWI data is from DATASET I, acquired with Neusoft 3T Universal scanner, acquisition matrix size 180×180, b value 1000 s/mm$^2$. (a-e) are from patient subject ID **PS#1,** 1$^{st}$ row and 2$^{nd}$ row of (f)-(j) are from patient subject ID **PS#2 and #3**, respectively.

Compared with multi-shot DWI reconstructed by the IRIS method[33] (**Fig. 6(d)**), which estimates inter-shot phase from navigators in an extra scan, the LoSP-Prompt (**Fig. 6(e)**) still has the advantage of improved SNR and reduced motion artifacts sourced from the spleen, although our method is navigator-free. Poor motion artifacts suppression of IRIS (arrows ② in **Fig. 6(d)**) may be due to the unreliable navigator echo because of low SNR or abdomen-introduced motion mismatch between navigator and image echoes.

For other two patients with abdominal lesions, including liver cysts (1st row in **Fig. 6(f-j)**) and hepatocellular carcinoma (2nd row in **Fig. 6(f-j)**), LoSP-Prompt shows better noise suppression and motion artifacts removal than LLR and MUSSELS, and provides clearer lesions.

Thus, LoSP-Prompt provides lower image artifacts and higher lesion detectability, even better than navigator-based approach.

**Generalization to multi-organ DWI**

Here, without algorithm modification or network re-training, LoSP-Prompt is applied to multi-organ DWI, especially for organs that are different from the synthesized training data (axial abdomen dataset). The LoSP-Prompt is compared with LLR and MUSSELS in high-resolution DWI reconstructions of 7 body parts, including kidney, sacroiliac, pelvis, knee, liver, spinal cord, and tumor brain. For each body part, we collect DWI data from at least 3 subjects.

DWI images of all 7 parts reconstructed by LoSP-Prompt (**Fig. 7(a)**) achieve nice motion artifacts removal and noise suppression, while LLR and MUSSELS do not generalize well in reconstructions of different slices of same organs and different organs. For sacroiliac DWI (2nd row in **Fig. 7(a)**), which suffers from very low SNR due to a high-resolution acquisition (matrix size = 320×336), LLR can hardly remove motion artifacts and noise. MUSSELS shows unstable performance at different slices of the subject HS#17 (nice reconstruction results at the 12th slice but fails on the 6th slice). More reconstructions of multi-body parts further show that MUSSELS are hardly to provide stable reconstructions, such as knee and spinal cord (4th and 5th rows in **Fig. 7(a)**). Given only a set of optimized subject-specific parameters, both LLR and MUSSELS are difficult to meet the needs of reconstructing DWI of heterogeneous organs in multi-slice or in multi-body parts.

Image quality is evaluated by 11 radiologists (with 6-, 10-, 11-, 12-, 15-, 16-, 20-, 22-, 25-, 27- and 32-year experiences) through independent and blind reader study[34]. In total, 188 slices are randomly selected from all reconstructions. Except for the spinal cord, liver, and tumor brain, about 30 slices are selected for each body part. For spinal cord, 18 slices from three subjects are employed. For liver, 50 slices from five subjects are selected. For brain tumor, 53 slices containing lesions are selected from 18 subjects. For each slice, 3 radiologists give independent scores in terms of three clinical criteria: SNR, artifact suppression, and overall image quality. Each criterion's score is ranged from 0 to 5 with a precision of 0.1 (i.e., 0~1: Non-diagnostic; 1~2: Poor; 2~3: Adequate; 3~4: Good; 4~5: Excellent). Statistical difference is indicated by the Wilcoxon signed-rank test on scores ($p < 0.05$).

Radiologists' evaluation (**Fig. 7(b)**) suggests that, LoSP-Prompt achieves the overall quality of 4-5 point (excellent) on kidney DWI, 4 points (good to excellent) on liver, sacroiliac and spinal cord DWI, and 3-4 points (good) on knee and tumor brain. In the brain tumor, LoSP-Prompt have comparable scores than MUSSELS, and better scores than LLR. Except for the brain tumor, in all reconstructions, LoSP-Prompt achieves better scores than MUSSELS and LLR, and has significant difference in most body parts.

Besides, LoSP-Prompt keeps a good stability in scores across the DWI reconstructions of all body parts. For example, in sacroiliac reconstruction, the standard deviation of the overall quality of LoSP-Prompt (0.40) is much smaller than that of LLR (1.12) and MUSSELS (1.40). This comparison shows that, the image-specific parameters provided by LoSP-Prompt are more robust than subject-specific reconstruction by traditional algorithms, and are more suitable for the reconstruction of areas with strong heterogeneity such as the sacroiliac and liver.

Thus, in high-resolution DWI of 7 body parts, LoSP-Prompt achieves better and more stable reconstruction than the cutting-edge algorithms, and obtain the best radiologists' ratings, indicating its potential for clinical applications.

## Discussion

### Limitations of LoSP-Prompt

Two main limitations restrict the reconstruction performance of LoSP-Prompt. The first is its ability to handle the cross-slice motion since we only discussed the multi-slice 2D sequences. The second is the relatively longer reconstruction time. For example, for a 2-shot liver DWI image with 2× uniform under-sampling (readout × phase encoding = 180×180, single average and diffusion direction), LoSP-Prompt costs 161.7 seconds, when using MATLAB on a server equipped with Intel Xeon Silver 4210 CPU and 256 GB RAM, although the prediction time of Prompt-Net is very short (0.9 seconds).

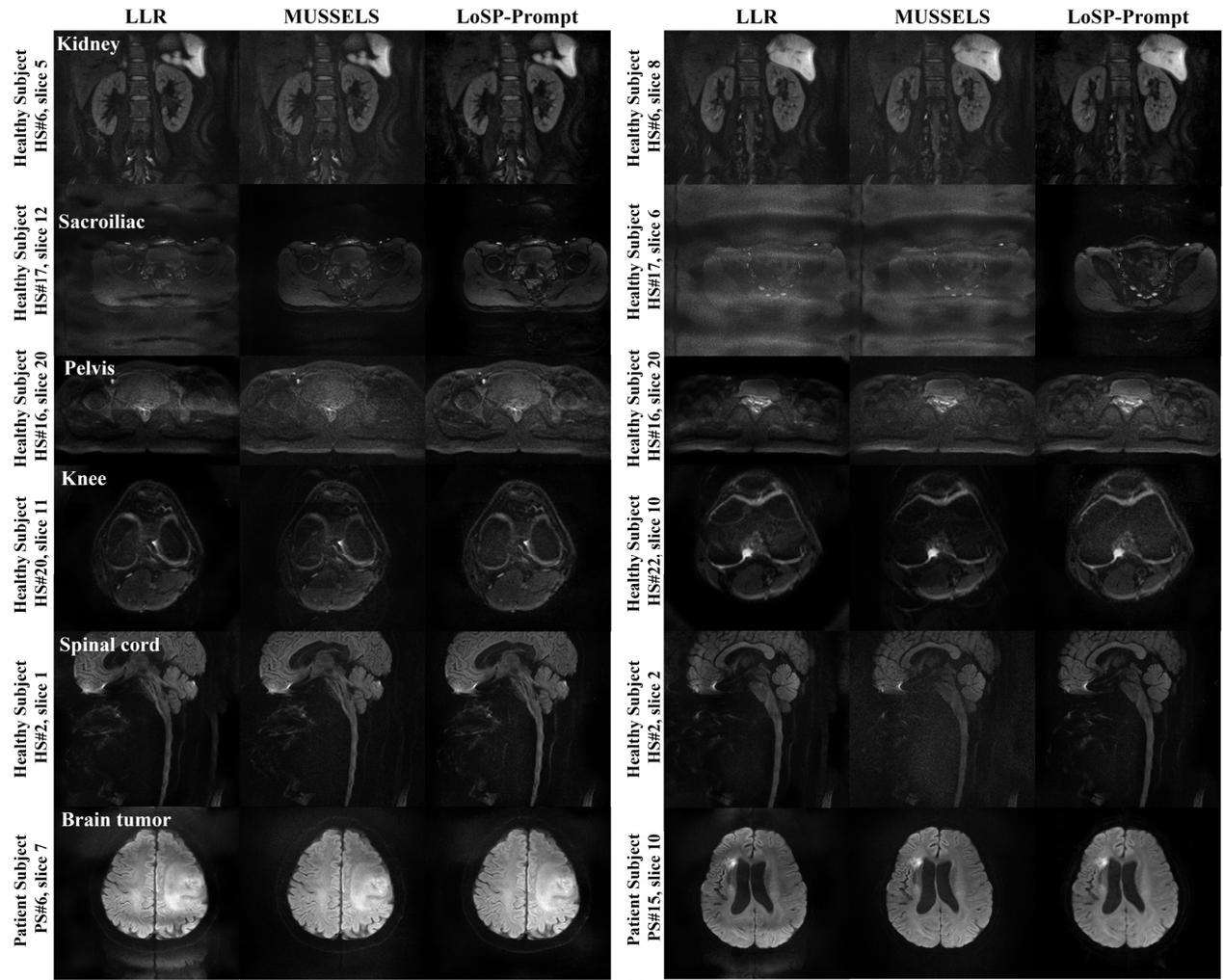
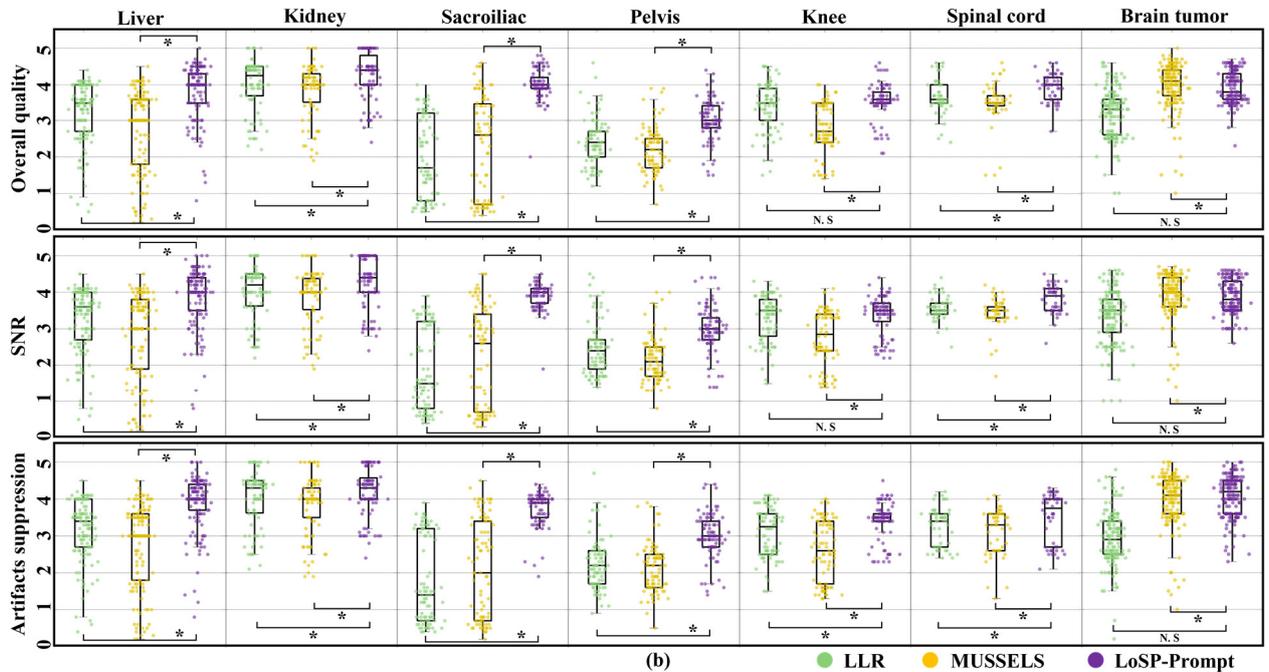

**Fig. 7 | Generalization to multi-organs. (a)** are reconstructions of kidney, sacroiliac, pelvis, knee, spinal cord, and tumor brain, by LLR, MUSSELS, and LoSP-Prompt. **(b)** are the subjective scores of MUSSELS, LLR, and LoSP-Prompt in terms of SNR, artifacts suppression, and overall quality. Note: * represents the significant difference between two methods ($p < 0.05$), and N. S. means no significant difference. Detailed imaging parameters is provided in Appendix **Note 2**. DWI images are acquired on healthy subject **HS#1, 2, 6, 16, 17, 20, 22**, and patient subject **PS#6, 15**.

# Conclusion

We propose a robust high-resolution DWI image reconstruction method (LoSP-Prompt) for multi-organ magnetic resonance imaging. Through 11 radiologists' evaluations, this method achieves good quality of DWI images in 7 body parts (10000+ DWI images from 43 subjects, 4 MRI models, 5 centers). Evidence on clear liver lesions in DWI images is consistent to pathological examination. This work pioneers a new framework that enables robust magnetic resonance image reconstruction via prompt and synthetic data learning, which may transform body-wide tumor diagnosis.

# Methods

## Physics-informed abdomen ms-iEPI DWI data synthesis and training data generation

The whole process of the training data synthesis (**Fig. 8**) in LoSP-Prompt include 7 steps: 1) Obtain abdomen DWI magnitude images **m** with organ masks; 2) Synthesize motion-induced phase **P** with a locally smooth phase model expressed by low-order and high-order polynomials; 3) Multiply magnitude images with synthetic motion-induced phases to get multi-shot images **I** = **Pm**; 4) Transform multi-shot image **I** into k-space (noise-free k-space $\mathbf{X}_{GT}$) and add Gaussian noise in k-space (noisy image $\mathbf{X}_{Inp}$, SNR range of 1-15 dB); 5) Separating noise-free and noisy image into paired ground truth and noisy 1D signal. 6) Perform Hankel Singular Value Decomposition and Truncated signal recovery (HSVDT) on 1D noisy signal, and compute the peak signal-to-noise-ratio (PSNR) of recovered 1D signal to noise-free 1D signal. 7) Take the optimal number of saved singular values to achieve highest PSNR in HSVDT for each 1D noisy signal as the network training label.

The step 1) is based on a publicly abdominal phantom (https://github.com/SeiberlichLab/Abdominal_MR_Phantom) that has 64 axial slices and each slice comprises 10 respiratory phases to form a complete respiratory cycle. 640 magnitude images are obtained. These slices encompass 14 anatomical structures $\mathbf{m}^o$ and their corresponding masks $\mathcal{U}^o$, including the adrenal glands, liver, gallbladder, stomach, pancreas, spleen, colon, kidneys, ureters, arteries, veins, muscles, bones, fat, and skin. The magnitude image **m** is defined as follows:

$$\mathbf{m} = \sum_{o=1}^{O} \mathbf{m}^o, \quad (1)$$

where $\mathbf{m}^o$ denotes the magnitude value assigned to the $o^{th}$ anatomical structure.

In step 2), in consideration of the organ-specific movements according to the respiration and heartbeat (Appendix **Note 1**), we introduce a locally smooth phase model for abdomen DWI motion-induced phase synthesis:

$$\mathbf{P}_j(x, y) = \sum_{o=1}^{O} \mathcal{U}^o \exp\{i \cdot (\sum_{l=0}^{L_o} \sum_{k=0}^{l} (A_{lk}^o x^k y^{l-k}))\}, j = 1, 2, ..., J, \quad (2)$$

where $x$ and $y$ denote the spatial coordinates of the image, $L_o$ represents the order of the generated phase for the $o^{th}$ anatomical structure, $A_{lk}^o$ and $\mathcal{U}^o$ denotes the phase generation parameter and binary mask for the $o^{th}$ anatomical structure, respectively, $J$ is the number of shots. By empirically setting different $L_o$ and $A_{lk}^o$ for distinct anatomical structures, one can use $\mathbf{P}_j(x,y)$ to simulate organ-specific motion-induced phases in realistic abdomen DWI. In total, 640 motion-induced phases matched with magnitude images are synthesized and we simulate $J = 2, 3, 4$ shots.

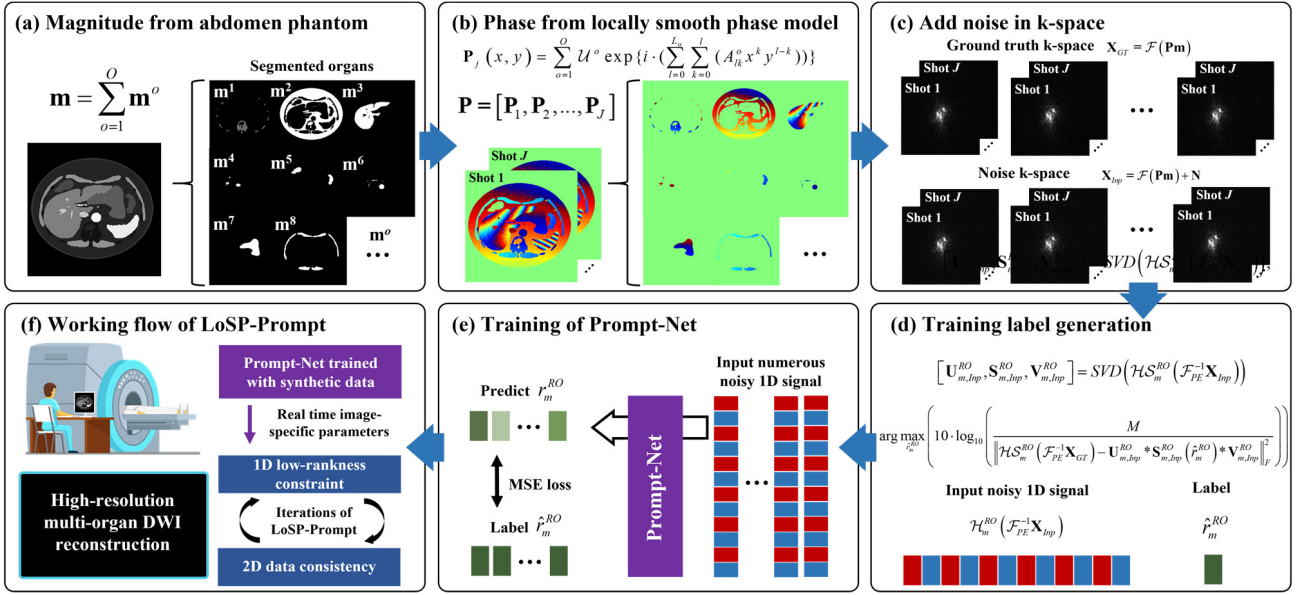

**Fig. 8 | Prompt learning from synthesized data provides LoSP reconstruction with image-specific parameters, enabling high-resolution multi-organ DWI.** (a) Magnitudes of adomenal organ phantom $\mathbf{m}^o$ and corresponding masks $\mathcal{U}^o$. (b) Organ-specific shot phase $\mathbf{P}_j$ generated from the locally smooth phase model. (c) Ground truth k-space $\mathbf{X}_{GT}$ and corresponding nosiy k-space $\mathbf{X}_{Inp}$ with added guassian noise **N**. (d) Generation of input noisy 1D signal and calculation of corresponding label. (e) The training of Prompt-Net with 100,000 pairs of syntesized samples. (f) The prompt learning boosts 1D low-rank reconstruction with image-specific parameters. Note: $A^{lm}$ is the effecients for phase generation of the $o^{th}$ organ. $\mathcal{F}$ is the 2D Fourier transform. $SVD$ is the singular value decomposition, $M$, $N$ are the dimentions of readout and phase encoding. $\mathcal{H}$ is the operation for 1D Hankel matrix construction. $\mathcal{S}_m^{RO}$ is the operation for selecting the $m^{th}$ readut line. $\mathcal{F}$ is the 1D inverse Fourier transofrm along phase encoding. MSE is the mean squared errror.

In step 5), the separated noisy 1D signals are $\mathcal{S}_m^{RO}(\mathcal{F}_{PE}^{-1}\mathbf{X}_{Inp})$ and $\mathcal{S}_n^{PE}(\mathcal{F}_{RO}^{-1}\mathbf{X}_{Inp})$, where $\mathcal{S}_m^{RO}$ and $\mathcal{S}_n^{PE}$ are the operator for extracting the $m^{th}$ readout line (frequency encoding) and $n^{th}$ phase encoding line (**Fig. S3** in **Appendix Note 3**), respectively; $\mathcal{F}_{RO}^{-1}$ and $\mathcal{F}_{PE}^{-1}$ are the 1D inverse Fourier transform along readout and phase encoding, respectively.

In step 6), the HSVDT on 1D noisy signal, and the PSNR of recovered 1D signal to noise-free 1D signal are as follows (take one readout line as an example), respectively:

$$\left[\mathbf{U}_{m,Inp}^{RO}, \mathbf{S}_{m,Inp}^{RO}, \mathbf{V}_{m,Inp}^{RO}\right] = SVD\left(\mathcal{H}\mathcal{S}_m^{RO}\left(\mathcal{F}_{PE}^{-1}\mathbf{X}_{Inp}\right)\right), \quad (3)$$

$$PSNR(r_m^{RO}) = 10*\log_{10}\left(\frac{M}{\left\|\mathcal{H}\mathcal{S}_m^{RO}\left(\mathcal{F}_{PE}^{-1}\mathbf{X}_{GT}\right) - \mathbf{U}_{m,Inp}^{RO} * \mathbf{S}_{m,Inp}^{RO}(r_m^{RO}) * \mathbf{V}_{m,Inp}^{RO}\right\|_F^2}\right), \quad (4)$$

where the $r_m^{RO}$ is the number of saved ranks; $SVD$ is the operator for singular value decomposition; $\mathbf{U}_{m,Inp}^{RO}, \mathbf{S}_{m,Inp}^{RO}, \mathbf{V}_{m,Inp}^{RO}$ are the left singular vectors, singular values, and right singular vectors, respectively; $m$ and $M$ are the index and total number of readout lines, respectively; $\mathbf{X}_{GT}$ is the k-space of corresponding noise-free data. $\mathcal{H}$ is an operator for constructing Hankel matrix[35] (**Fig. S3** in Appendix **Note 3**).

In step 7), the optimal number of saved singular values is found by solving the problem:

$$\hat{r}_m^{RO} = \arg\min_{r_m^{RO}} \frac{1}{PSNR(r_m^{RO})}, \quad r_m^{RO} \in \mathbb{N}^+. \quad (5)$$

Since $\hat{r}_m^{RO} \in \mathbb{N}^+$, the most suitable solution for Eq. (5) can be found by a traversal way.

Following steps 1) - 7), we finally generate $640 \times 256 \times 2$ ($M$ and $N$ = 256) paired of 1D noisy input signal $\mathcal{S}_m^{RO}(\mathcal{F}_{PE}^{-1}\mathbf{X}_{Inp})$ (or $\mathcal{S}_n^{PE}(\mathcal{F}_{RO}^{-1}\mathbf{X}_{Inp})$) and parameter label $\hat{r}_m^{RO}$ (or $\hat{r}_n^{PE}$) in total.

### Training of Prompt-Net with synthesized data

To maintain generality, a modified ResNet18 network[26] (Appendix **Note 4**) is selected as the basic network architecture of the Prompt-Net. The loss function of network training is defined as the mean square error between the predicted saved ranks and their labels $\hat{r}_m^{RO}$ (or $\hat{r}_n^{PE}$). Then, the well-trained Prompt-Net could automatically provide image-specific reconstruction parameters $r_m^{RO}$ (or $r_n^{PE}$) in the multi-shot DWI reconstructions.

### DWI reconstruction model with Locally Smooth Phase prior (LoSP) and prompt learning (LoSP-Prompt)

The proposed basic 1D reconstruction model, LoSP, is:

$$\min_{\mathbf{X}} \frac{\lambda}{2}\left\|\mathbf{Y} - \mathcal{UFC}\mathcal{F}^{-1}\mathbf{X}\right\|_F^2 + \sum_{m=1}^{M}\left\|\mathcal{HS}_m^{RO}(\mathcal{F}_{PE}^{-1}\mathbf{X})\right\|_* + \sum_{n=1}^{N}\left\|\mathcal{HS}_n^{PE}(\mathcal{F}_{RO}^{-1}\mathbf{X})\right\|_*, (6)$$

where $\mathbf{Y}$ denotes the acquired multi-shot, multi-coil k-space data; $\mathcal{U}$ represents the sampling mask corresponding to the multi-shot data; $\mathcal{F}$ and $\mathcal{F}^{-1}$ represent 2D Fourier transform and its inverse, respectively; $\mathbf{C}$ is the coil sensitivity maps; $\mathbf{X}$ is the k-space of the target multi-shot DWI image; $\|\cdot\|_*$ and $\|\cdot\|_F$ are the nuclear norm and Frobenius norm, respectively; $\lambda$ is the regularization parameter, and is set to 1 by default.

The LoSP in Eq. (6) is solved by an Alternating Direction Method of Multipliers algorithm (Appendix **Note 3**). In the solving process of LoSP, minimizing the nuclear norm of Hankel matrix lifted from each 1D signal is optimized by singular value truncation with a fixed truncation parameter $r$ (the number of saved singular values, i.e. saved rank).

LoSP is enhanced to LoSP-Prompt by replacing fixed $r$ with 1D signal-specific truncation parameters $r_m^{RO}$ ( or $r_n^{PE}$ ), that are automatically predicted by Prompt-Net (Appendix **Note 3**).

### Compared methods and evaluation criteria

The proposed LoSP-Prompt (and its baseline LoSP) are compared with 7 reconstruction methods, namely IRIS[33], MUSE[7], MUSSELS[16], S-LORAKS[17], PAIR[18], LLR[29], and DONATE[19]. Reasons for selecting these algorithms include: IRIS is a navigator-based method, which employs the navigator echo (in additional scan) for motion-induced phase correction; MUSE is a classic method with multiplexed coil sensitivity coding; MUSSELS, S-LORAKS, and PAIR are all cutting-edge 2D low-rank reconstruction methods, and they all assume that the motion-induced phase has global smooth characteristics; LLR is a constrained reconstruction using the locally low-rankness in the image domain introduced by the local smooth phase; DONATE is a separated 1D low-rank reconstruction in the readout (frequency coding) direction, and the low-rank prior used is based on the 1D compact support in the image domain, which differs from high-order phase modeling proposed in this work. MUSE and DONATE are compared in the appendix.

Among these methods, codes of MUSSELS, DONATE, LLR, and PAIR are provided by original authors; IRIS[33], MUSE[7] and S-LORAKS[17] are reproduced according to the corresponding papers. In all experiments, all methods are given with a set of optimized subject-specific parameters (not image-specific) to best balance motion artifacts removal and noise suppression. That means, each algorithm's parameters are optimized for this for each subject but not specific b-value/slice/direction/average DWI image (Appendix **Note 9**).

PSNR is used as an objective indicator to evaluate the reconstruction performance in simulated study (**Fig. 2**). Three clinical-concerned subjective metrics are adopted in the reader study, including the SNR, artifact suppression, and overall quality of reconstructed DWI images. The reader study is performed through the cloud computing platform, CloudBrain-ReconAI[34], which is free to access at https://csrc.xmu.edu.cn/CloudBrain.html and has been used multiple work[19,36].

**Data preprocessing and postprocessing**

Two DWI databases (Healthy and patient subjects) are collected and all experiments are Institutional Review Board-approved. Before reconstructions, the echo-planar imaging ghost is corrected.

For reconstructions, coil sensitivity maps are estimated by ESPIRIT[37] with non-diffusion (b-value = 0) data or pre-scan data. The images of multiple shots are combined and displayed by taking the square root of the square sum. ADC maps are estimated by the least square method.

**Data and code availability**

The training data and code will be shared respectively at:

https://github.com/qianchne/LoSP-Prompt

**Acknowledgements**

The authors thank Mathews Jacob, Yuxin Hu, and Justin Haldar for sharing their codes online, Hui Zhang and Chengyan Wang for helpful discussion, Runhan Chen, Lin Guo, and Jiaxin Zhou for assisting data collection. This work was supported in part by the National Natural Science Foundation of China under grants 62331021, 62371410 and 62122064, Natural Science Foundation of Fujian Province of China under grants 2023J02005, Industry-university Cooperation Projects of the Ministry of Education of China under grant 231107173160805, National Key Research and Development Program of China under grant 2023YFF0714200, Zhou Yongtang Fund for High Talents Team under grant 0621-Z0332004, President Fund of Xiamen University under grant 20720220063, and Nanqiang Outstanding Talent Program of Xiamen University.

**Author contributions**

X. Qu and C. Qian conceived the idea and designed the experiments, X. Qu supervised the project, C. Qian implemented the method, C. Qian, H. Zhang, and Q. Cai processed databases, and produced results, C. Qian, and H. Zhang drew the figures for the manuscript and supplementary. J. Ma, L. Zhu, Y. Wang, R. Song, L. Li, L. Mei, X. Jiang, B. Jiang, R. Tao, C. Chen, S. Chen, H. Zhong, J. Zhou, and Q. Xu helped to acquire *in vivo* data. C. Qian, H. Zhang, L. Zhu, X. Huang, D. Liang, H. Wang, Q. Guo, J. Lin, T. Kang, M. Lu, L. Fu, R. Huang, and J. Wang conducted the reader study. The manuscript was drafted by C. Qian and improved by D. Guo, and X. Qu. X. Qu and D. Guo acquired research funds and provided all the needed resources.

**Competing interests**

Y. Wang, R. Song, L, Li, L. Mei, X. Jiang, and Q. Xu are the employees of Neusoft Medical Systems. B. Jiang and R. Tao are the employees of United Imaging Healthcare. X. Qu received research grants from Neusoft Medical Systems.

# Supplementary Information

## Note 1. Motion model and phase model

Assume that the coordinates of any point in the imaging target coordinate system (also the image coordinate system) are $\vec{r}$, and the coordinate system of the magnetic resonance gradient field is $\vec{R}$. Then the position of any point in the imaging target at time $t$ in the magnetic resonance gradient field coordinate system is $\vec{R} = \vec{r} + \Delta\vec{R}(\vec{r},t)$. The signal obtained at this time is:

$$\begin{aligned}\mathbf{Y}_h\left(\vec{k}(t)\right) &= \int \mathbf{C}_h(\vec{r}) e^{-i2\pi\vec{k}(t)\vec{R}} \mathbf{m}(\vec{r}) d^3\vec{r} = \int \mathbf{C}_h(\vec{r}) e^{-i2\pi\vec{k}(t)(\vec{r}+\Delta\vec{R}(\vec{r},t))} \mathbf{m}(\vec{r}) d^3\vec{r} \\ &= \int \mathbf{C}_h(\vec{r}) e^{-i2\pi\vec{k}(t)\Delta\vec{R}(\vec{r},t)} \mathbf{m}(\vec{r}) e^{-i2\pi\vec{k}(t)\vec{r}} d^3\vec{r} = \int \mathbf{C}_h(\vec{r}) e^{-i\Delta\phi(\vec{r},t)} \mathbf{m}(\vec{r}) e^{-i2\pi\vec{k}(t)\vec{r}} d^3\vec{r}.\end{aligned} \quad (S.1)$$

Therefore, the motion phase due to motion displacement is:

$$\Delta\phi(\vec{r},t) = 2\pi \cdot \vec{k}(t) \cdot \Delta\vec{R}(\vec{r},t). \quad (S.2)$$

Substituting formula (S.2) into the above formula (S.1), we can get the real-time motion phase change:

$$\Delta\phi(\vec{r},t) = \gamma \int_0^t \Delta\vec{R}(\vec{r},t') \cdot \vec{G}(t') dt'. \quad (S.3)$$

To simplify, we give the motion phase after the gradient field ends (T is the total time of the gradient field):

$$\Delta\phi(\vec{r}) = \gamma \int_0^T \Delta\vec{R}(\vec{r},t) \cdot \vec{G}(t) dt. \quad (S.4)$$

The motion displacement $\Delta\vec{R}(\vec{r},t)$ determines the nature of the motion phase (**TABLE I**). During the entire imaging process (after the gradient field ends), when no motion occurs ($\vec{R} = \vec{r}$, $\Delta\vec{R}(\vec{r},t) = 0$), $\Delta\phi = 0$; when only translational motion occurs ($\vec{R} = \vec{r} + \Delta R$, $\Delta R$ is a constant independent of position $\vec{r}$), $\Delta\phi = \Delta R \cdot \gamma \int_0^T \vec{G}(t) dt$, is also a constant with position $\vec{r}$, which is called the 0-order motion phase; when only rigid body rotation occurs, the rotation angle is $\theta$ ($\vec{R} = \vec{r} + \theta \cdot \vec{r}$), the motion phase is a function that changes linearly with position $\vec{r}$, which is called the 1-order motion phase; when non-rigid body motion occurs, the motion displacement becomes more complex and becomes a high-order function of $\vec{r}$ ($\vec{R} = \vec{r} + \sum_{l=0}^{L} \alpha_l \vec{r}^l$, L represents the order), $\Delta\phi(\vec{r}) = \sum_{l=0}^{L} \alpha_l \vec{r}^l \cdot \gamma \int_0^T \vec{G}(t) dt$, also becomes a high-order function of $\vec{r}$, which is called the high-order motion phase; when the motion displacement becomes more complex and becomes a piecewise time-sharing function of $\vec{r}$ and $t$ (for example, different non-rigid body motions occur in organs in different regions), $\Delta\phi(\vec{r})$ becomes more complex and cannot be directly expressed by a combination of high-order terms of $\vec{r}$, but its local motion phase may still conform to the high-order function form of $\vec{r}$, which inspires us to use polynomial functions to approximate, fit, and generate the organ-specific motion phases (**Fig. S1(a)**).

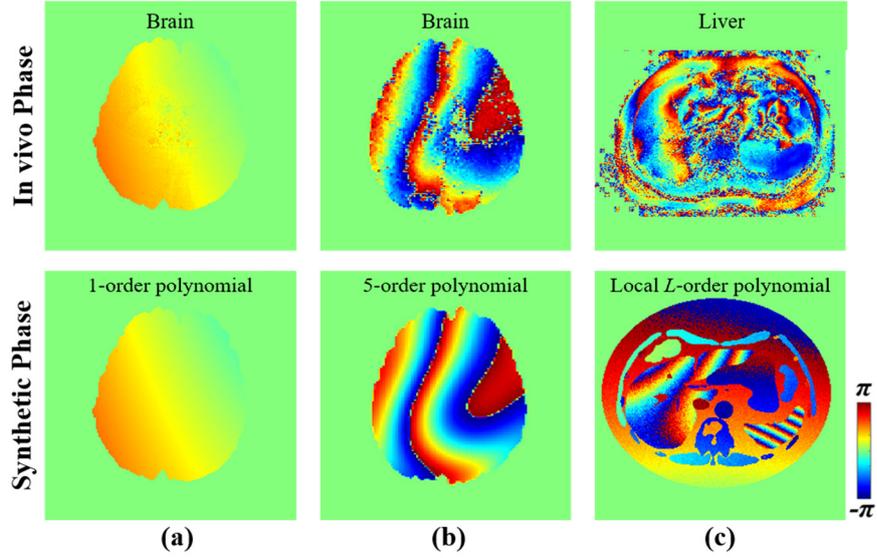

**Figure S1: Representative motion phase models.** (a) is the motion phase in brain, which conforms to low-order smooth modeling and can be fitted using a 1-order polynomial model. (b) is the motion phase in brain, which conforms to high-order smooth modeling and can be fitted using a 5-order polynomial model. (c) is the motion phase in upper abdomen, which conforms to the local smooth characteristics and can be simulated using a local *L*-order polynomial model to synthesize the organ-specific phase. *L* is different for distinct organs.

**TABLE I. Qualitative analysis of motion displacement $\Delta \vec{R}(\vec{r},t)$ and motion phase**

| Type of movement displacement | Motion phase | Motion phase model |
|---|---|---|
| No movement $\Delta \vec{R}(\vec{r},t) = 0$ | No motion phase $\Delta\phi = 0$ | $\Delta\phi(x,y) = 0$ |
| Rigid body translation $\Delta \vec{R}(\vec{r},t) = \Delta R$ | 0-order motion phase (Low-order smoothness) $\Delta\phi = \Delta R \cdot \phi_G$ | Constant $\Delta\phi(x,y) = A_{00}$ |
| Rigid body rotation $\Delta \vec{R}(\vec{r},t) = \theta \cdot \vec{r}$ | 1-order motion phase (Low-order smoothness) $\Delta\phi(\vec{r}) = \theta \cdot \vec{r} \cdot \phi_G$ | 1-order polynomial $\Delta\phi(x,y) = A_{01}x + A_{10}y$ |
| Non-rigid *L*-order motion/deformation $\Delta \vec{R}(\vec{r},t) = \sum_{l=0}^{L} \alpha_l \vec{r}^l$ | *L*-order motion phase (high-order smoothness) $\Delta\phi(\vec{r}) = \sum_{l=0}^{L} \alpha_l \vec{r}^l \cdot \phi_G$ | *L*-order polynomial $\Delta\phi(x,y) = \sum_{l=0}^{L} \sum_{m=0}^{l} (A_{lm} x^m y^{l-m})$ |
| Approximate non-rigid *L*-order motion/deformation (Organ-specific motions) $\Delta \vec{R}(\vec{r},t) \approx \sum_{o=1}^{O} \mathcal{U}^o \left( \sum_{l=0}^{L_o} \alpha_l \vec{r}^l \right)$ | Local *L*-order motion phase (Organ-specific) $\Delta\phi(\vec{r}) \approx \sum_{o=1}^{O} \mathcal{U}^o \left( \sum_{l=0}^{L_o} \alpha_l \vec{r}^l \right) \cdot \phi_G$ | local *L*-order polynomial $\Delta\phi(x,y) \approx \sum_{o=1}^{O} \mathcal{U}^o \left( \sum_{l=0}^{L} \sum_{m=0}^{l} (A_{lm} x^m y^{l-m}) \right)$ |

Note: $\phi_G = \gamma \int_0^T \vec{G}(t) dt$, $\mathcal{U}^o$ is the $o^{th}$ local region, $A_{lm}$ is the coefficient in the polynomial phase model.

Base on the above analysis, we define the 0- and 1- order motion phases introduced by rigid motion as low-order smooth phases, which are suitable for typical imaging scenarios such as the brain (**Fig. S1(a)**); we define the $L \geq 2$-order phases introduced by non-rigid motion/deformation as high-order smooth phases, which are suitable for typical imaging scenarios such as the brain and spinal cord under the pulsation of cerebrospinal fluid (**Fig. S1(b)**); we define the regional approximate high-order phases introduced by the regional approximate non-rigid $L$-order motion/deformation as organ-specific phases, which are suitable for typical imaging scenarios such as the abdomen under the influence of breathing, heart, intestinal peristalsis and other movements (**Fig. S1(c)**).

In the previous work, the motion phase assumption based on low-order and high-order smoothness is widely exploited in high-resolution DWI reconstructions. A series of state-of-the-art (SOTA) reconstruction methods are designed with the assumption that, the motion phase is smooth or composed of smooth functions, such as ALOHA[1], MUSSELS[2,3], LORAKS[4,5], PAIR[6], DONATE[7], and MoDL-MUSSELS[8], bring promising results in high-solution brain DWI reconstructions. However, none of these methods analyzes and utilizes the motion phase properties of positions with intense movements, such as the abdomen.

In this work, we use a local $L$-order polynomial model for organ-specific motion phase synthesis in the training data generation.

## Note 2. Scan parameters

### TABLE II. Scan parameters for DATASET I (Patient DWI)

| Sequence | Multi-shot Interleaved Echo Planar Imaging DWI Sequence | |
|---|---|---|
| Body parts | Abdomen | Brain |
| Plane | Axial | Axial |
| Scanner | Neusoft, 3.0T, Universal | Philips, 3.0T, Ingenia CX |
| Shot number | 2 | 4 |
| Coil number | 28 | 16 |
| Acceleration / Sampling rate | SENSE / 0.5, PF/ 0.9 | / |
| TR/TE (ms) | Trigger / 63.5 | 3000 / 90 |
| b-value (number of averages) (s/mm$^2$) | 50 (1) <br> 1000 (3) | 0 (1) <br> 1000 (1) |
| Diffusion directions | 3 | 12 |
| Matrix size (RO×PE) | 180×180 | 180×180 |
| FOV (mm) | 360×360 | 240×240 |
| Slice | 30 | 12 / 24 |
| Slice thickness (mm) | 5 | 5 |
| Subjects | 3 | 18 |
| Subject ID | PS#1-3 | PS#4-21 |
| Scan time (min: sec) | ~3:45 | 5:48 |

Note: SENSE means the uniform under-sampling (**Fig. S2(b)**); PF is the partial Fourier under-sampling (**Fig. S2(c)**); TR is the time of repetition; TE is the time of echo; RO (readout) and PE (phase encoding) represent frequency and phase encoding, respectively; FOV is the field of view. PS is the patient subject.

**Figure S2: Schematic diagram of sampling mask (take 4-shot as an example).** (a-c) are the sampling mask of fully sampling, uniform under-sampling and partial Fourier under-sampling, respectively. Note: The solid line of each color represents the sampled lines of a shot, and the dashed line represents unsampled data. The sampling rate is calculated by $\frac{Sampled\ lines}{Total\ lines}$, and marked above the sampling mask.

TABLE III. Scan parameters for DATASET II (Healthy volunteer DWI)

| Sequence | Multi-shot Interleaved Echo Planar Imaging DWI Sequence | | | | | | | |
|---|---|---|---|---|---|---|---|---|
| Part | Brain & neck | Upper abdomen | | | | Lower abdomen | | Limbs |
| Target organ / Plane | Spinal cord / Sagittal | Kidney / Coronal | Liver / Axial | | | Uterus / Axial | Sacroiliac / Axial | Knee / Axial |
| Scanner | Philips, 3T, Ingenia CX | NeuMR, 3T, Universal | NeuMR, 3T, Universal | | UI, 5T, Jupiter | NeuMR, 3T, Universal | UI, 3T uMR 890 | NeuMR, 3T Universal |
| Shot number | 4 | 2 | 2 | 2 | 4 | 2 | 4 | 2 |
| Coil number | 16 | 28 | 26 | 26 | 26 | 24 | 26 | 24 |
| Acceleration /Sampling rate | PF / 0.6 | SENSE / 0.5 PF / 0.9 | SENSE / 0.5 PF / 0.9 | SENSE / 0.5 PF / 0.9 | / | SENSE / 0.5 | PF / 0.83 | SENSE / 0.5 |
| TR/TE (ms) | 1411/53 | Trigger / 63 | Trigger / 63 | Trigger / 61 | / | 4000/69 | 4245/63.5 | 5000/61.5 |
| b-value (number of averages) (s/mm$^2$) | 0 (2) 1000 (4) | 50 (1) 800 (3) | 50 (1) 1000 (3) | 50 (1) 1000 (3) | 50 (1) 800 (2) | 50 (2) 1000 (3) 1500 (4) | 50 (1) 600 (2) | 0 (1) 500 (2) |
| Diffusion directions | 3 | 3 | 3 | 3 | 3 | 3 | 3 | 3 |
| Matrix size (RO×PE) | 252×156 | 256×256 | 256×256 | 180*144 | 256×256 | 180×180 | 320×336 | 160×160 |
| FOV (mm) | 300×240 | 340×340 | 340×340 | 340×340 | 380×380 | 340×340 | 400×400 | 160×160 |
| Slice | 10 | 12 | 34 | 34 | 34 | 25 | 18 | 20 |
| Slice thickness (mm) | 5 | 4 | 4 | 4 | 5 | 4 | 4 | 3 |
| Subjects | 3 | 4 | 2 | 3 | 1 | 3 | 3 | 3 |
| Subject ID | HS#1-3 | HS#4-7 | HS#8-9 | HS#10-12 | HS#13 | HS#14-16 | HS#17-19 | HS#20-22 |
| Scan time (min: sec) | 01:23 | ~ 03:13 | ~ 03:06 | ~ 03:06 | / | 03:43 | 02:17 | 1:20 |

Note: SENSE means the uniform under-sampling (**Fig. S2(b)**); PF is the partial Fourier under-sampling (**Fig. S2(c)**); TR is the time of repetition; TE is the time of echo; RO (readout) and PE (phase encoding) represent frequency and phase encoding, respectively; FOV is the field of view. HS is the healthy subject.

## TABLE VI. Information of patient subject

| Subject ID | Disease information | Traceable clinical evidence | Data Usage |
|---|---|---|---|
| PS#1 | Primary hepatocellular carcinoma | Pathology of liver puncture biopsy<br>Diagnosis of MRI | Fig. 7 |
| PS#2 | Pancreatic tail cancer, liver cyst | Diagnosis of MRI | Fig. 7 |
| PS#3 | Primary hepatocellular carcinoma | Diagnosis of MRI | Fig. 7 |
| PS#4 | Brain metastasis from small cell carcinoma of the thoracic esophagus | Pathology of gastric tube mucosal biopsy,<br>Diagnosis of brain MRI | Fig. 8 |
| PS#5 | Brain metastasis from gastroesophageal junction adenocarcinoma | Pathology of lymph node puncture biopsy<br>Diagnosis of MRI and CT | Fig. 8 |
| PS#6 | Secondary epilepsy, brain metastasis from lung adenocarcinoma, and cerebral ischemia | Pathology of lung puncture biopsy,<br>Diagnosis of MRI and CT | Fig. 8 |
| PS#7 | Brain metastasis from lung adenocarcinoma | Postoperative pathology<br>Diagnosis of brain MRI | Fig. 8 |
| PS#8 | Brain metastasis from small cell lung cancer, cerebral ischemia | Pathology of lung puncture biopsy<br>Diagnosis of brain MRI | Fig. 8 |
| PS#9 | Brain metastasis from small cell lung cancer, cerebral ischemia | Pathology of lung puncture biopsy<br>Diagnosis of brain MRI | Fig. 8 |
| PS#10 | Brain metastasis from lung adenocarcinoma, cerebral hemorrhage, cerebral ischemia | Pathology of lung puncture biopsy<br>Diagnosis of brain MRI | Fig. 8 |
| PS#11 | Secondary epilepsy, brain metastasis from lung adenocarcinoma | Pathology of lung puncture biopsy<br>Diagnosis of brain MRI | Fig. 8 |
| PS#12 | Brain metastasis from lung adenocarcinoma | Postoperative pathology<br>Diagnosis of brain MRI | Fig. 8 |
| PS#13 | Brain metastasis from lung adenocarcinoma | Pathology of lung puncture biopsy<br>Diagnosis of brain MRI | Fig. 8 |
| PS#14 | Brain metastasis from lung adenocarcinoma | Pathology of lung puncture biopsy<br>Diagnosis of brain MRI | Fig. 8 |
| PS#15 | Brain metastasis from breast cancer | Postoperative pathology<br>Diagnosis of brain MRI | Fig. 8 |
| PS#16 | Brain metastasis from lung adenocarcinoma, cerebral ischemia | Postoperative pathology<br>Diagnosis of brain MRI | Fig. 8 |
| PS#17 | Brain metastasis from small cell lung cancer | Pathology of lung puncture biopsy<br>Diagnosis of brain MRI | Fig. 8 |
| PS#18 | Brain metastasis from small cell lung cancer | Pathology of lung puncture biopsy<br>Diagnosis of brain MRI and PET/CT | Fig. 8 |
| PS#19 | Brain metastasis from lung adenocarcinoma | Pathology of lung puncture biopsy<br>Diagnosis of brain MRI | Fig. 8 |
| PS#20 | Brain metastasis from lung adenocarcinoma | Diagnosis of brain MRI | Fig. 8 |
| PS#21 | Brain metastasis from lung adenocarcinoma, cerebral ischemia | Pathology of lung puncture biopsy<br>Diagnosis of brain MRI | Fig. 8 |

Note: PS is the patient subject. MRI is Magnetic Resonance Imaging. CT is Computational Tomography. PET is Positive Emission Tomography.

## Note 3. Numerical algorithm of LoSP and LoSP-prompt

The proposed LoSP has the cost function as follows:

$$\min_{\mathbf{X}} \frac{\lambda}{2}\|\mathbf{Y} - \mathcal{U}\mathcal{F}\mathbf{C}\mathcal{F}^{-1}\mathbf{X}\|_F^2 + \sum_{m=1}^{M}\|\mathcal{H}\mathcal{S}_m^{RO}\left(\mathcal{F}_{PE}^{-1}\mathbf{X}\right)\|_* + \sum_{n=1}^{N}\|\mathcal{H}\mathcal{S}_n^{PE}\left(\mathcal{F}_{RO}^{-1}\mathbf{X}\right)\|_*, \tag{S.5}$$

where $\mathbf{Y}$ denotes the acquired multi-shot, multi-coil k-space data, $\mathcal{U}$ represents the sampling mask corresponding to the multi-shot data, $\mathcal{F}$ and $\mathcal{F}^{-1}$ represent 2D Fourier transform and its inverse, respectively, $\mathbf{C}$ is the coil sensitivity maps, $\mathbf{X}$ is the k-space of the target multi-shot DWI image, $\|\cdot\|_*$ and $\|\cdot\|_F$ are the nuclear norm and Frobenius norm, respectively, $\lambda$ is the regularization parameter, and is set to 1 by default, $\mathcal{S}_m^{RO}$ and $\mathcal{S}_n^{PE}$ are the operator for extracting the $m^{th}$ readout line and $n^{th}$ phase encoding line, respectively, $\mathcal{F}_{PE}^{-1}$ and $\mathcal{F}_{RO}^{-1}$ are the 1D Fourier transform along phase encoding and readout, respectively, $M$ and $N$ are the number of readout and phase encoding lines, respectively, $\mathcal{H}$ is an operator for constructing Hankel matrix[9] (**Fig. S3**).

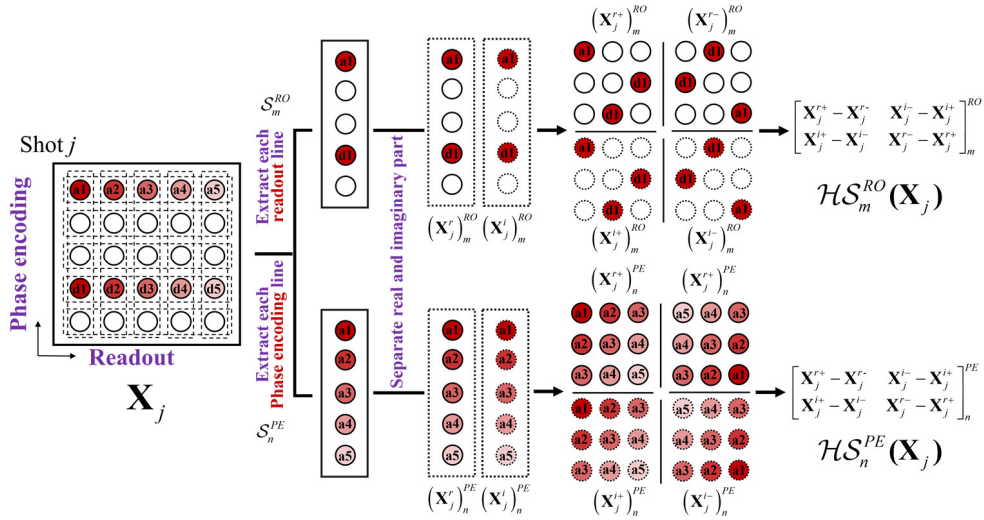

**Figure S3: Schematic diagram of the operator** $\mathcal{S}_m^{RO}, \mathcal{S}_n^{PE}, \mathcal{H}$ **in LoSP model**. The superscripts $r$ and $i$ denote the real and imaginary components, respectively. Note: For simplicity, only the construction process for the $j^{th}$ shot is shown, and the 1D Fourier transform is omitted in the schematic.

To minimize the cost function, the Alternating Direction Method of Multipliers (ADMM) algorithm is derived to fit our problem.

First, the auxiliary variables are defined as follows:

$$\mathbf{Z}_m^{RO} = \mathcal{H}\mathcal{S}_m^{RO}\left(\mathcal{F}_{PE}^{-1}\mathbf{X}\right), \quad m = 1, 2, ..., M, \tag{S.6}$$

$$\mathbf{Z}_n^{PE} = \mathcal{H}\mathcal{S}_n^{PE}\left(\mathcal{F}_{RO}^{-1}\mathbf{X}\right), \quad n = 1, 2, ..., N. \tag{S.7}$$

Thus, the augmented Lagrangian form of the cost function becomes:

$$\max_{\mathbf{D}_m^{RO},\mathbf{D}_n^{PE}} \min_{\mathbf{X},\mathbf{Z}_m^{RO},\mathbf{Z}_n^{PE}} \frac{\lambda}{2}\|\mathbf{Y}-\mathcal{UFC}\mathcal{F}^{-1}\mathbf{X}\|_F^2$$
$$+\sum_{m=1}^{M}\left(\|\mathbf{Z}_m^{RO}\|_* + <\mathbf{D}_m^{RO},\mathcal{HS}_m^{RO}\left(\mathcal{F}_{PE}^{-1}\mathbf{X}\right)-\mathbf{Z}_m^{RO}> + \frac{\rho}{2}\|\mathcal{HS}_m^{RO}\left(\mathcal{F}_{PE}^{-1}\mathbf{X}\right)-\mathbf{Z}_m^{RO}\|_F^2\right) \quad \text{(S.8)}$$
$$+\sum_{n=1}^{N}\left(\|\mathbf{Z}_n^{PE}\|_* + <\mathbf{D}_n^{PE},\mathcal{HS}_n^{PE}\left(\mathcal{F}_{RO}^{-1}\mathbf{X}\right)-\mathbf{Z}_n^{PE}> + \frac{\rho}{2}\|\mathcal{HS}_n^{PE}\left(\mathcal{F}_{RO}^{-1}\mathbf{X}\right)-\mathbf{Z}_n^{PE}\|_F^2\right),$$

where $\mathbf{D}_m^{RO}$ and $\mathbf{D}_n^{PE}$ are the Lagrange multipliers, $<\cdot,\cdot>$ represents the inner product of the complex matrix in the Hilbert space, $\rho$ is the penalty parameter. By exchanging the maximum and minimum, we get the dual model of the above model:

$$\min_{\mathbf{X},\mathbf{Z}_m^{RO},\mathbf{Z}_n^{PE}} \max_{\mathbf{D}_m^{RO},\mathbf{D}_n^{PE}} \frac{\lambda}{2}\|\mathbf{Y}-\mathcal{UFC}\mathcal{F}^{-1}\mathbf{X}\|_F^2$$
$$+\sum_{m=1}^{M}\left(\|\mathbf{Z}_m^{RO}\|_* + <\mathbf{D}_m^{RO},\mathcal{HS}_m^{RO}\left(\mathcal{F}_{PE}^{-1}\mathbf{X}\right)-\mathbf{Z}_m^{RO}> + \frac{\rho}{2}\|\mathcal{HS}_m^{RO}\left(\mathcal{F}_{PE}^{-1}\mathbf{X}\right)-\mathbf{Z}_m^{RO}\|_F^2\right) \quad \text{(S.9)}$$
$$+\sum_{n=1}^{N}\left(\|\mathbf{Z}_n^{PE}\|_* + <\mathbf{D}_n^{PE},\mathcal{HS}_n^{PE}\left(\mathcal{F}_{RO}^{-1}\mathbf{X}\right)-\mathbf{Z}_n^{PE}> + \frac{\rho}{2}\|\mathcal{HS}_n^{PE}\left(\mathcal{F}_{RO}^{-1}\mathbf{X}\right)-\mathbf{Z}_n^{PE}\|_F^2\right).$$

This optimization problem can be divided into five sub-problems, and the final result is obtained by alternatively and iteratively solving each sub-problem:

$$\mathbf{Z}_m^{RO(k+1)} = \arg\min_{\mathbf{Z}_m^{RO(k+1)}} \sum_{m=1}^{M}\|\mathbf{Z}_m^{RO(k+1)}\|_* + \frac{\rho}{2}\|\mathcal{HS}_m^{RO}\left(\mathcal{F}_{PE}^{-1}\mathbf{X}^{(k)}\right)-\mathbf{Z}_m^{RO(k+1)} + \frac{\mathbf{D}_m^{RO(k)}}{\rho}\|_F^2, \quad \text{(S.10)}$$

$$\mathbf{Z}_n^{PE(k+1)} = \arg\min_{\mathbf{Z}_n^{PE(k+1)}} \sum_{n=1}^{N}\|\mathbf{Z}_m^{PE(k+1)}\|_* + \frac{\rho}{2}\|\mathcal{HS}_n^{PE}\left(\mathcal{F}_{RO}^{-1}\mathbf{X}^{(k)}\right)-\mathbf{Z}_n^{PE(k+1)} + \frac{\mathbf{D}_n^{PE(k)}}{\rho}\|_F^2, \quad \text{(S.11)}$$

$$\mathbf{X}^{k+1} = \arg\min_{\mathbf{X}^{k+1}} \frac{\lambda}{2}\|\mathbf{Y}-\mathcal{UFC}\mathcal{F}^{-1}\mathbf{X}^{k+1}\|_F^2 + \sum_{m=1}^{M}\left(\frac{\rho}{2}\|\mathcal{HS}_m^{RO}\left(\mathcal{F}_{PE}^{-1}\mathbf{X}^{k+1}\right)-\mathbf{Z}_m^{RO(k+1)} + \frac{\mathbf{D}_m^{RO}}{\rho}\|_F^2\right)$$
$$+ \sum_{n=1}^{N}\left(\frac{\rho}{2}\|\mathcal{HS}_n^{PE}\left(\mathcal{F}_{RO}^{-1}\mathbf{X}^{k+1}\right)-\mathbf{Z}_n^{PE(k+1)}\|_F^2 + \frac{\mathbf{D}_n^{PE}}{\rho}\|_F^2\right), \quad \text{(S.12)}$$

$$\mathbf{D}_m^{(k+1)} = \mathbf{D}_m^{(k)} + \tau(\mathcal{HS}_m^{RO}\mathcal{F}_{PE}^{-1}\mathbf{X}^{(k+1)}-\mathbf{Z}_m^{(k+1)}), \quad m=1,2,...,M. \quad \text{(S.13)}$$

$$\mathbf{D}_n^{(k+1)} = \mathbf{D}_n^{(k)} + \tau(\mathcal{HS}_n^{PE}\mathcal{F}_{RO}^{-1}\mathbf{X}^{(k+1)}-\mathbf{Z}_n^{(k+1)}), \quad n=1,2,...,N. \quad \text{(S.14)}$$

where $\tau$ is the step size. Subproblems (S.10) and (S.11) can be solved by singular value truncation:

$$\mathbf{Z}_m^{RO(k+1)} = \mathbf{S}_r(\mathcal{HS}_m^{RO}\left(\mathcal{F}_{PE}^{-1}\mathbf{X}^{(k)}\right)+\frac{\mathbf{D}_m^{RO(k)}}{\rho}), \quad \text{(S.15)}$$

$$\mathbf{Z}_n^{PE(k+1)} = \mathbf{S}_r(\mathcal{HS}_n^{PE}\left(\mathcal{F}_{RO}^{-1}\mathbf{X}^{(k)}\right)+\frac{\mathbf{D}_n^{PE(k)}}{\rho}), \quad \text{(S.16)}$$

where $\mathbf{S}_r(\mathbf{A})$ represents the singular value decomposition of the matrix $\mathbf{A}$, and its singular value truncation, retaining the first $r$ largest singular values and the corresponding subspace to restore the signal.

Subproblem (S.12) can be solved using the following closed-form solution:

$$\mathbf{X}^{(k+1)} = (\lambda \mathcal{F} \mathbf{C}^* \mathcal{F}^{-1} \mathcal{U}^* \mathcal{U} \mathcal{F} \mathbf{C} \mathcal{F}^{-1} + \rho \sum_{m=1}^{M} \mathcal{F}_{PE} \mathcal{S}_m^{RO*} \mathcal{H}^* \mathcal{H} \mathcal{S}_m^{RO} \mathcal{F}_{PE}^{-1} + \rho \sum_{n=1}^{N} \mathcal{F}_{RO} \mathcal{S}_n^{PE*} \mathcal{H}^* \mathcal{H} \mathcal{S}_n^{PE} \mathcal{F}_{RO}^{-1})^{-1}$$

$$(\lambda \mathcal{F} \mathbf{C}^* \mathcal{F}^{-1} \mathcal{U}^* \mathbf{Y} + \sum_{m=1}^{M} \rho \mathcal{F}_{PE} \mathcal{S}_m^{RO*} \mathcal{H}^* \mathbf{Z}_m^{RO(k+1)} - \mathcal{F}_{PE} \mathcal{S}_m^{RO*} \mathcal{H}^* \mathbf{D}_m^{RO(k)}$$  (S.17)

$$+ \sum_{n=1}^{N} \rho \mathcal{F}_{RO} \mathcal{S}_n^{PE*} \mathcal{H}^* \mathbf{Z}_n^{PE(k+1)} - \mathcal{F}_{RO} \mathcal{S}_n^{PE*} \mathcal{H}^* \mathbf{D}_n^{PE(k)}).$$

where $\mathcal{U}^*, \mathcal{S}_m^{RO*}, \mathcal{S}_n^{PE*}, \mathcal{H}^*, \mathbf{C}^*$ are the adjoint operators, $\mathcal{F}_{RO}, \mathcal{F}_{PE}$ are the 1D Fourier transforms of readout and phase encoding, respectively.

LoSP-Prompt enhances the reconstruction performance of LoSP through signal-specific automatically saved ranks $r_n^{RO}$ (and $r_n^{PE}$) provided by Prompt-Net. In the process of solving subproblems (S.10) and (S.11), Prompt-Net is used to predict the optimal singular value truncation parameters as follows:

$$r_n^{RO} = PromptN\left(\mathcal{F}_{PE} \mathcal{S}_m^{RO*} \mathcal{H}^*\left(\mathcal{H}_n^{RO}\left(\mathcal{F}_{PE}^{-1} \mathbf{X}^{(k)}\right) + \frac{\mathbf{D}_n^{RO(k)}}{\rho}\right), \hat{\Theta}\right),$$  (S.18)

$$r_n^{PE} = PromptN\left(\mathcal{F}_{RO} \mathcal{S}_n^{PE*} \mathcal{H}^*\left(\mathcal{H}_n^{PE}\left(\mathcal{F}_{RO}^{-1} \mathbf{X}^{(k)}\right) + \frac{\mathbf{D}_n^{PE(k)}}{\rho}\right), \hat{\Theta}\right),$$  (S.19)

$$\mathbf{Z}_m^{RO(k+1)} = \mathbf{S}_{r_n^{RO}}(\mathcal{H}\mathcal{S}_m^{RO}\left(\mathcal{F}_{PE}^{-1} \mathbf{X}^{(k)}\right) + \frac{\mathbf{D}_m^{RO(k)}}{\rho}), \quad \mathbf{Z}_n^{PE(k+1)} = \mathbf{S}_{r_n^{PE}}(\mathcal{H}\mathcal{S}_n^{PE}\left(\mathcal{F}_{RO}^{-1} \mathbf{X}^{(k)}\right) + \frac{\mathbf{D}_n^{PE(k)}}{\rho}),$$  (S.20)

where $PromptN(\cdot, \hat{\Theta})$ is the well-trained Prompt-Net with model weights $\hat{\Theta}$.

Compared with LoSP, the core improvement of LoSP-Prompt is to replace the fixed saved rank $r$ of all 1D signals with automatically saved ranks $r_n^{RO}$ (and $r_n^{PE}$), which is readout-specific, phase encoding-specific, and image-specific for each 1D signal. Thus, automatically saved ranks could ensure that 1D signals from different regions are subject to differentiated low-rank constraints, thereby achieving robust reconstruction in multi-b-value, multi-slice, multi-organ, healthy and patient DWI images.

## Note 4. Network architecture and training of Prompt-Net

This section introduces the architecture and training of Prompt-Net. To maintain generality, the most common Res-Net18 is selected as the basic network architecture of the Prompt-Net (**Fig. S4**). The main structure contains 16 convolutional layers. At the output end, three fully connected layers are added as network outputs.

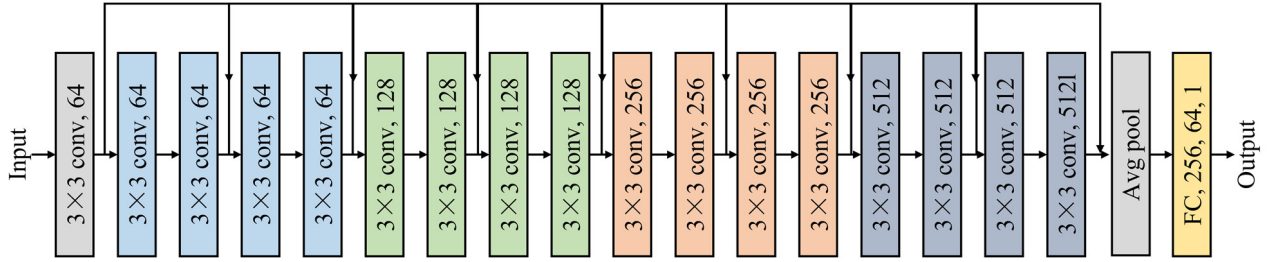

**Figure S4: Network architecture of Prompt-Net**. Note: Conv is a convolutional layer, Avg pool is an average pooling, FC is a three-layer fully connected layer, and each fully connected layer contains an activation function layer.

The loss function of its training is as follows:

$$\mathcal{L}(\hat{\Theta}) = \min_{\Theta} \sum_{t=1}^{T}\sum_{m=1}^{M} \left\| PromptN\left(\mathcal{S}_m^{RO}\left(\mathcal{F}_{PE}^{-1}\mathbf{X}_{Inp}^t\right), \Theta\right) - \hat{r}_n^{RO} \right\|_F^2 + \sum_{t=1}^{T}\sum_{n=1}^{N} \left\| PromptN\left(\mathcal{S}_n^{PE}\left(\mathcal{F}_{RO}^{-1}\mathbf{X}_{Inp}^t\right), \Theta\right) - \hat{r}_n^{PE} \right\|_F^2, \quad (S.21)$$

where $T$ is the total number of training samples. After training, Prompt-Net obtains the optimal weights $\hat{\Theta}$.

In the training data synthesis stage, we generate three datasets ($J$ = 2, 3, 4), and each dataset contains 32,7680 pairs of 1D synthetic data. For each dataset, 10,0000 pairs of data are randomly selected for training. 90% of them are used as training sets, and 10% are used as validation sets to adjust network hyperparameters and preliminarily evaluate the network's reconstruction performance.

In the training stage, the Adam optimizer is used for 100 rounds, with an initial learning rate set to 0.001 and decayed by a factor of 0.9 every 50 rounds, and a batch size of 256. The proposed basic network was developed based on the MATLAB Neural Network Toolkit and trained and run on a server equipped with an Intel Xeon Silver 4210 CPU (256 GB RAM memory) and an Nvidia Tesla T4 GPU (16 GB memory). The typical training time was about 6 hours.

In the reconstruction stage, for the given multi-shot multi-channel DWI k-space data $\mathbf{Y}$ and its corresponding channel sensitivity maps $\mathbf{C}$, the LoSP and LoSP-Prompt can be used for reconstructing DWI image. LoSP-Prompt costs 161.7 seconds for reconstructing a 2-shot liver DWI image with 2× uniform under-sampling (readout × phase encoding = 180×180), when using MATLAB on a server equipped with Intel Xeon Silver 4210 CPU and 256 GB RAM, although the prediction time of Prompt-Net is very short (less than 1 second) and can be ignored.

## Note 5. More results for high-resolution abdomen DWI reconstruction

This section gives all the reconstruction results in **Figs. 5-14**. In the reconstructions of the entire upper abdomen DWI (Subject ID is **HS#13**), each comparison method employs a set of optimized reconstruction parameters (subject-specific), while LoSP-Prompt can use automatically image-specific parameters provided by Prompt-Net.

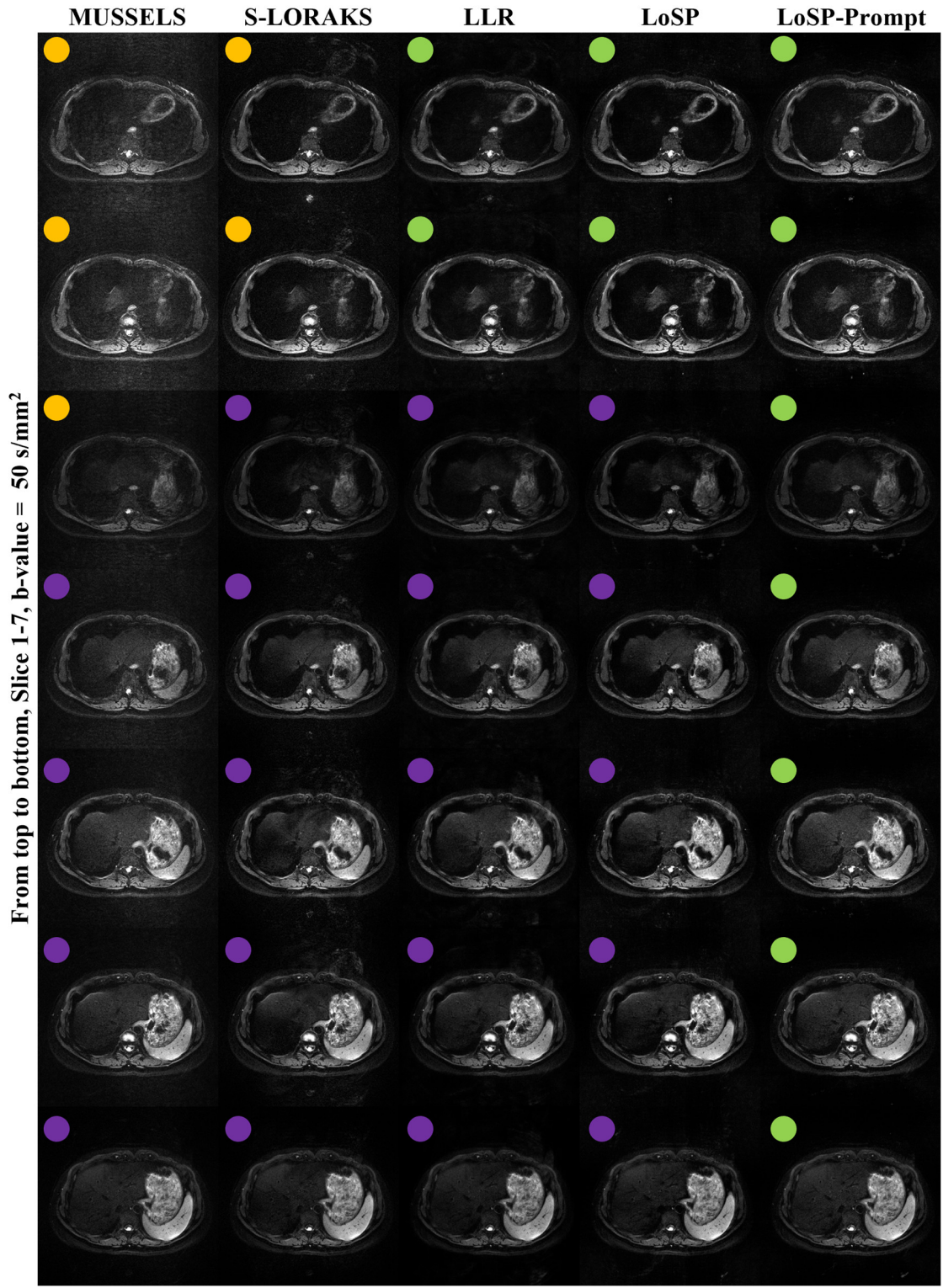

**Figure S5: Reconstructed DWI images of b-value 50 s/mm² (slice 1-7).** The purple, yellow, and green circular marks in the upper left corner indicate that the image has large noise/artifacts residual, structural signal loss, and good quality, respectively.

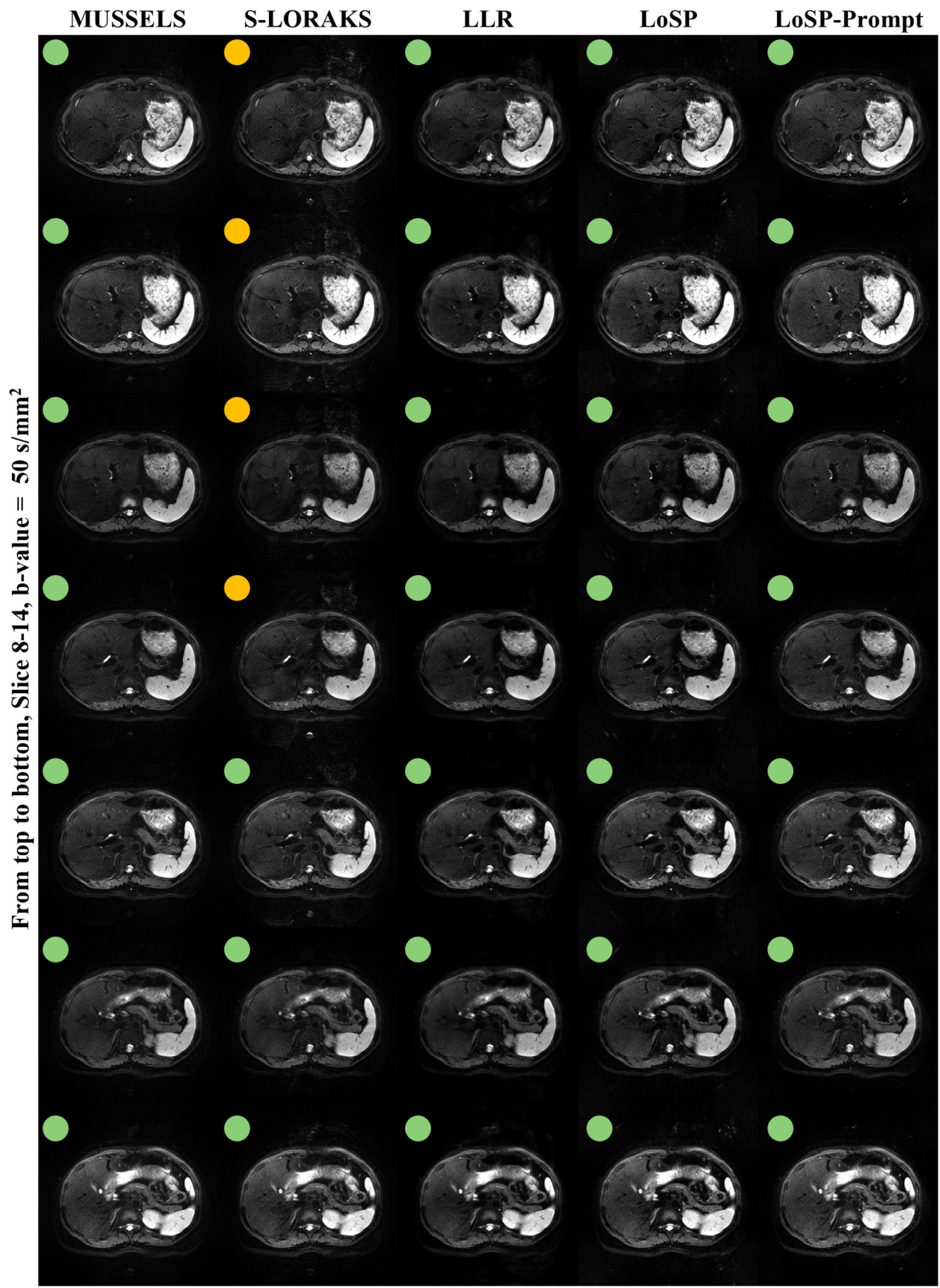

**Figure S6: Reconstructed DWI images of b-value 50 s/mm² (slice 8-14).** The purple, yellow, and green circular marks in the upper left corner indicate that the image has large noise/artifacts residual, structural signal loss, and good quality, respectively.

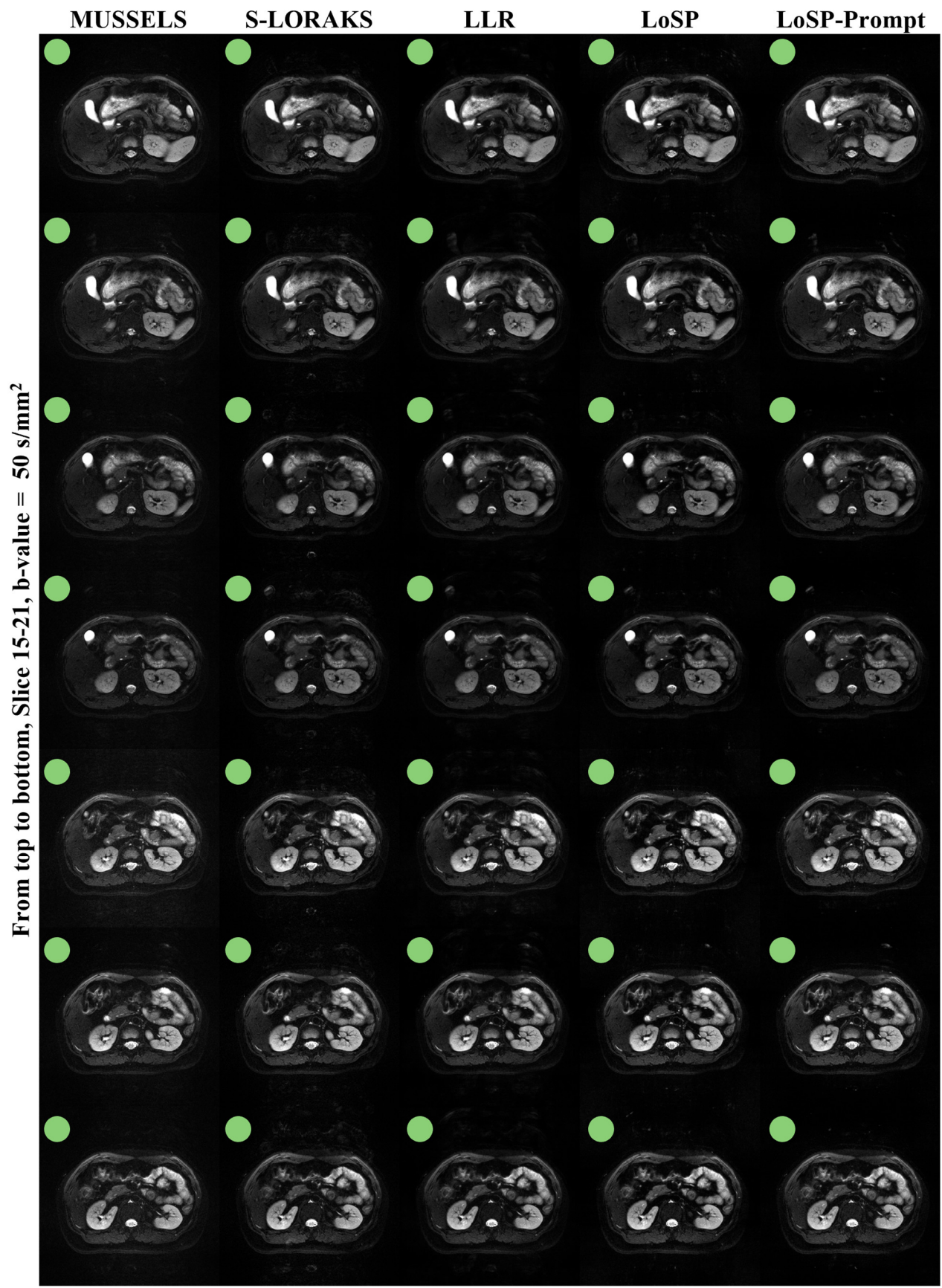

**Figure S7: Reconstructed DWI images of b-value 50 s/mm² (slice 15-21).** The purple, yellow, and green circular marks in the upper left corner indicate that the image has large noise/artifacts residual, structural signal loss, and good quality, respectively.

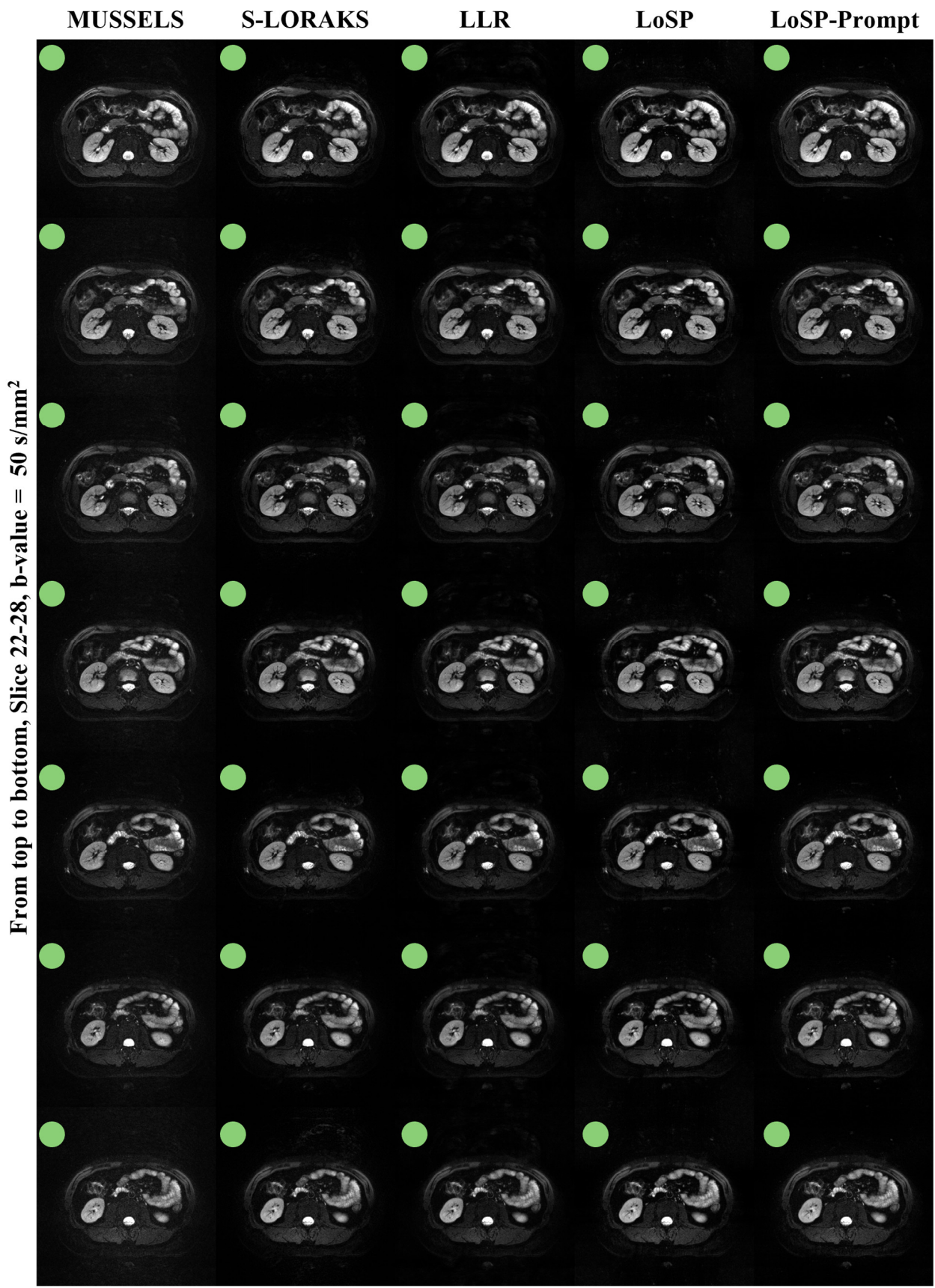

**Figure S8: Reconstructed DWI images of b-value 50 s/mm² (slice 22-28).** The purple, yellow, and green circular marks in the upper left corner indicate that the image has large noise/artifacts residual, structural signal loss, and good quality, respectively.

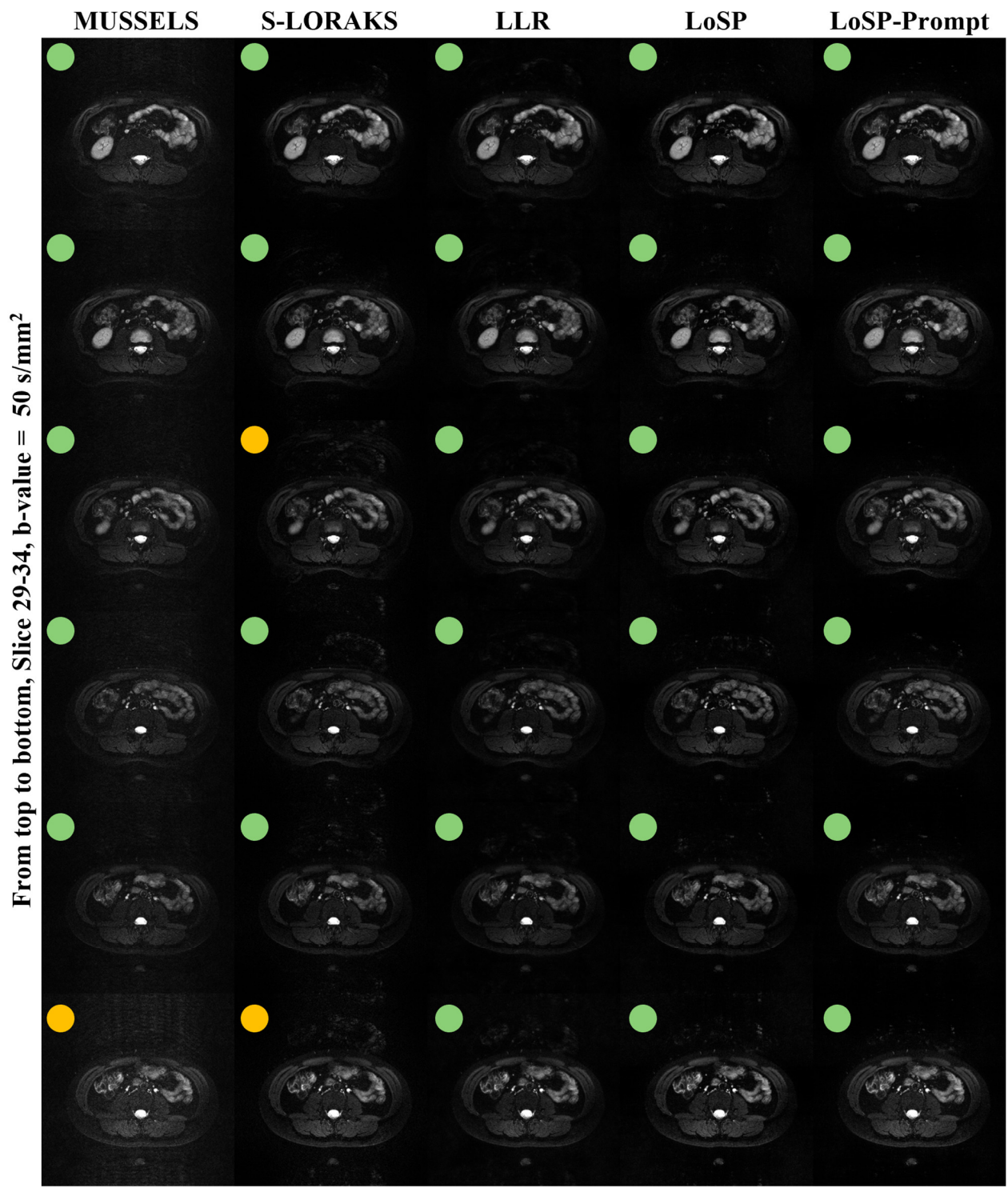

**Figure S9: Reconstructed DWI images of b-value 50 s/mm² (slice 29-34).** The purple, yellow, and green circular marks in the upper left corner indicate that the image has large noise/artifacts residual, structural signal loss, and good quality, respectively.

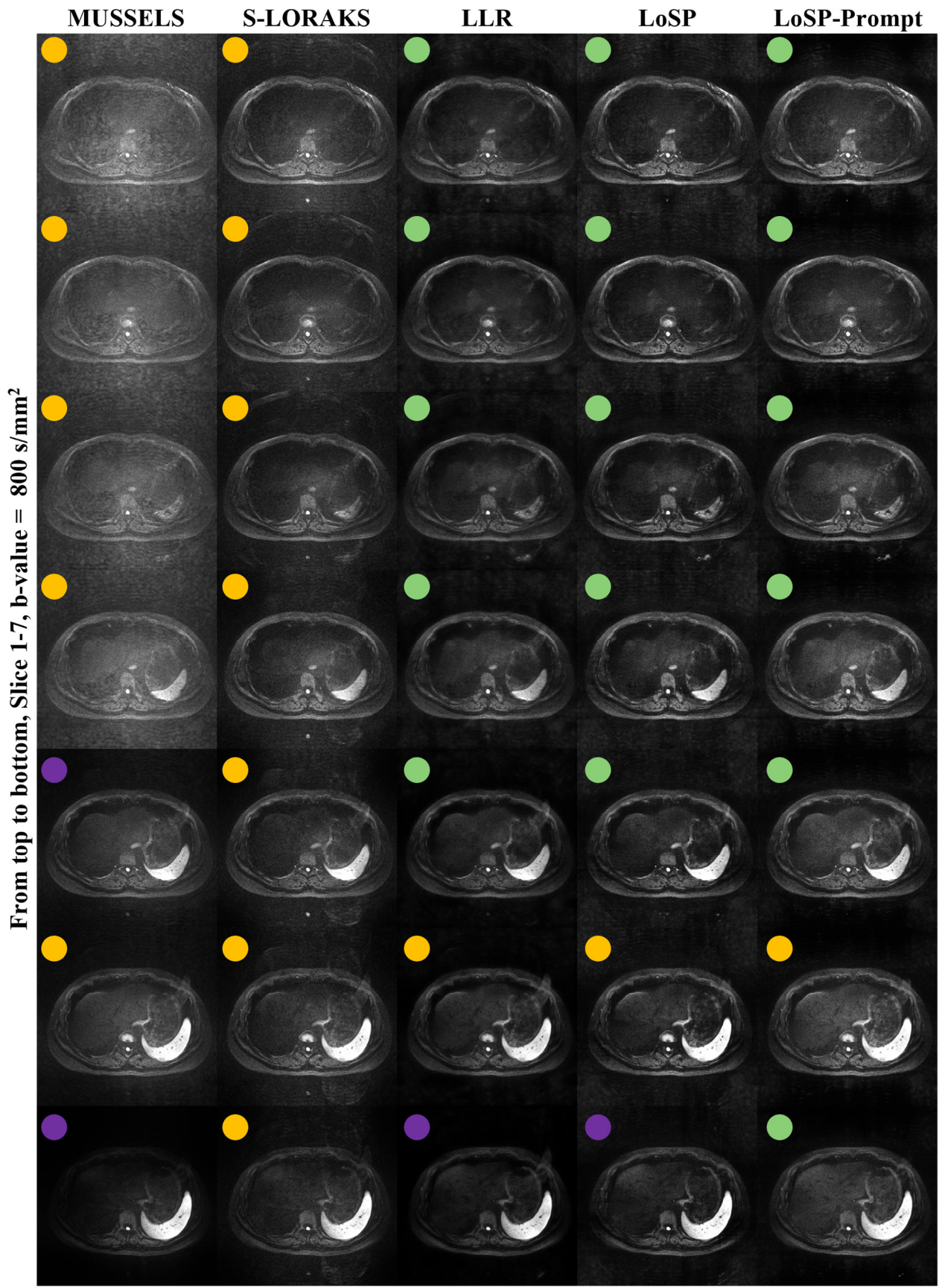

**Figure S10: Reconstructed DWI images of b-value 800 s/mm² (slice 1-7).** The purple, yellow, and green circular marks in the upper left corner indicate that the image has large noise/artifacts residual, structural signal loss, and good quality, respectively.

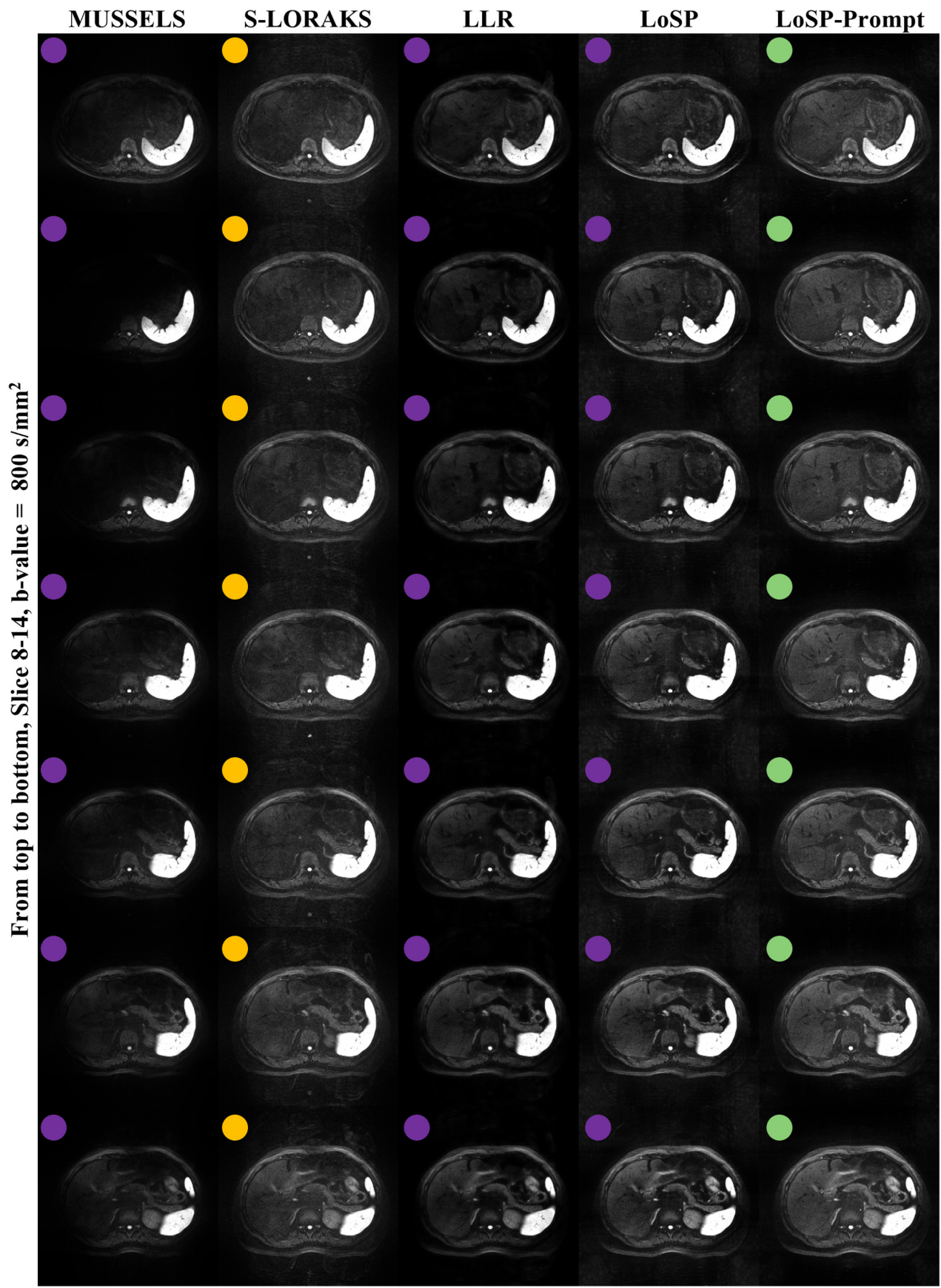

**Figure S11: Reconstructed DWI images of b-value 800 s/mm² (slice 8-14).** The purple, yellow, and green circular marks in the upper left corner indicate that the image has large noise/artifacts residual, structural signal loss, and good quality, respectively.

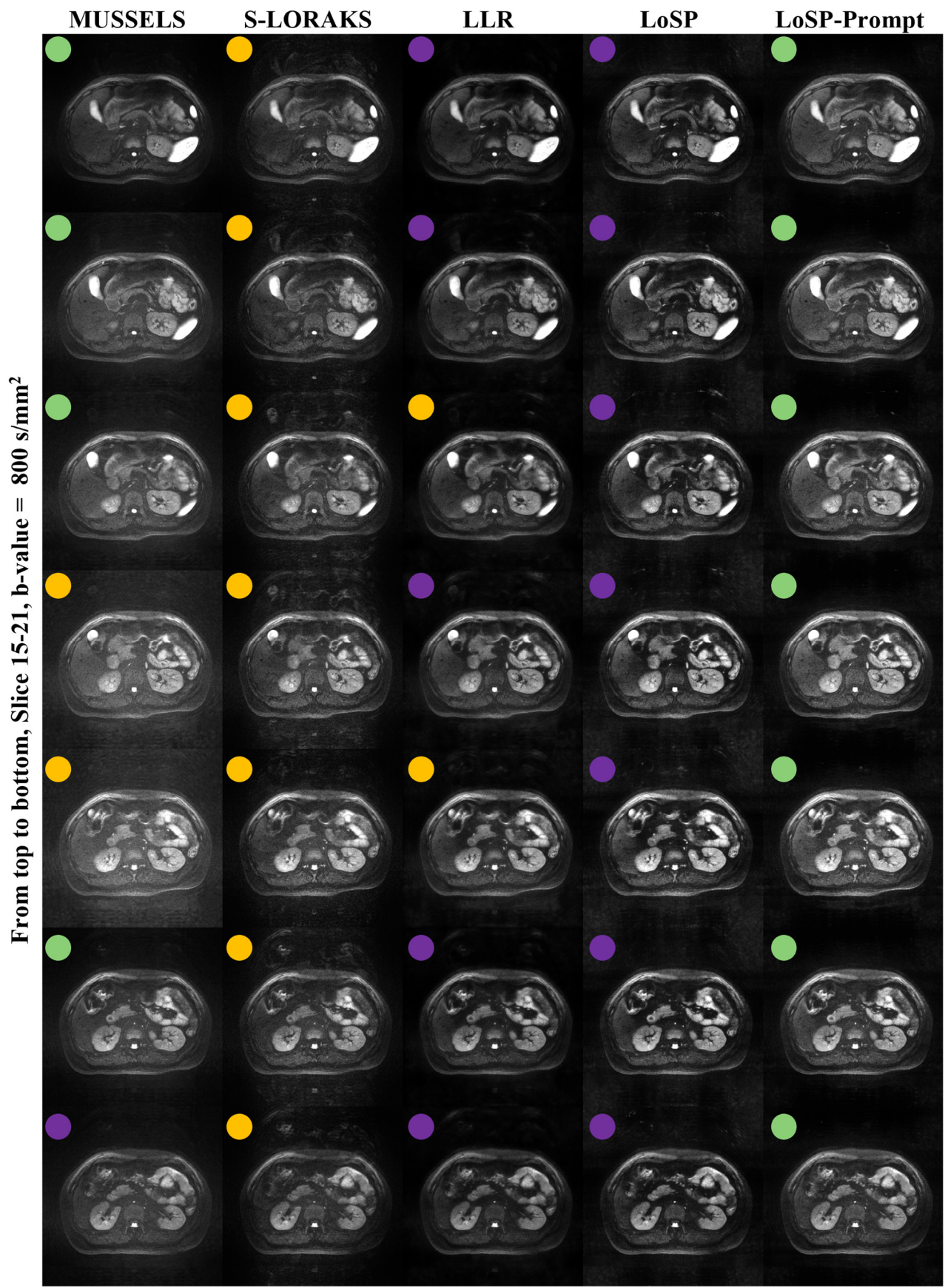

**Figure S12: Reconstructed DWI images of b-value 800 s/mm² (slice 15-21).** The purple, yellow, and green circular marks in the upper left corner indicate that the image has large noise/artifacts residual, structural signal loss, and good quality, respectively.

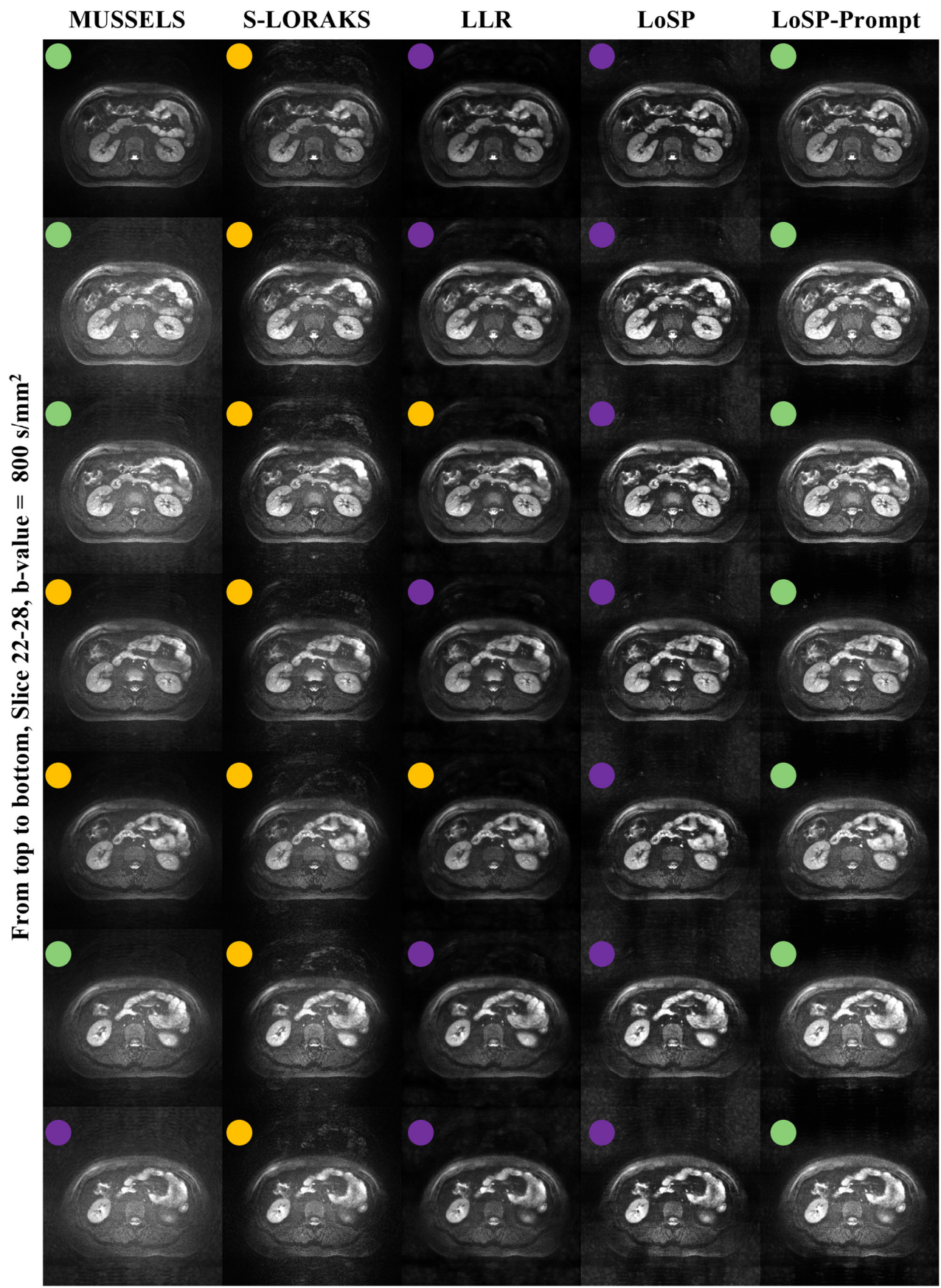

**Figure S13: Reconstructed DWI images of b-value 800 s/mm² (slice 22-28).** The purple, yellow, and green circular marks in the upper left corner indicate that the image has large noise/artifacts residual, structural signal loss, and good quality, respectively.

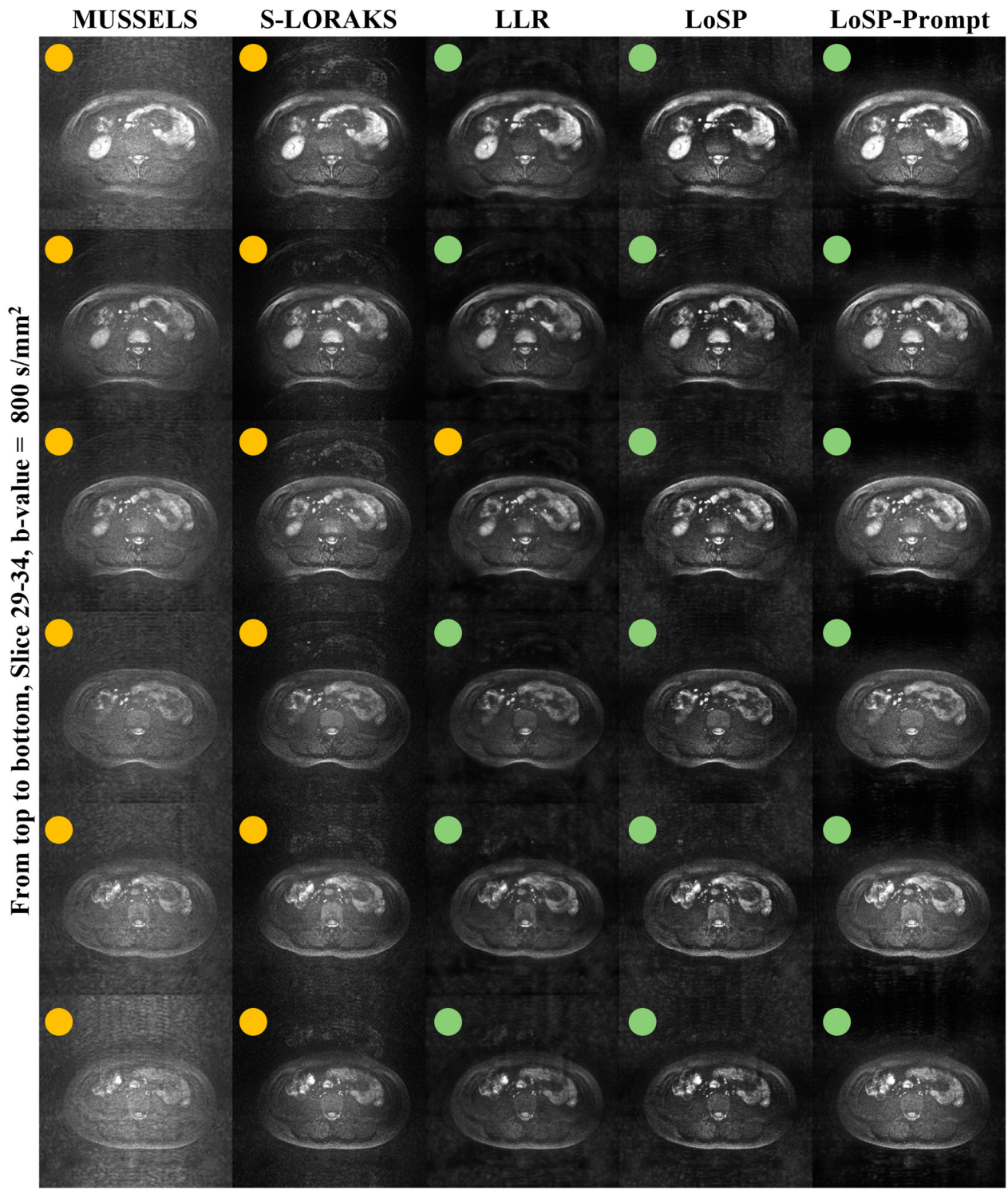

**Figure S14: Reconstructed DWI images of b-value 800 s/mm² (slice 29-34).** The purple, yellow, and green circular marks in the upper left corner indicate that the image has large noise/artifacts residual, structural signal loss, and good quality, respectively.

## Note 6. Comparison study on accelerated reconstructions with SOTA methods

The proposed methods are compared with 6 state-of-the-art (SOTA) methods on two acceleration scenarios: 2× uniform under-sampling (sampling rate = 0.5) and partial Fourier under-sampling (sampling rate = 0.7).

First, in the reconstruction comparison of 2× uniform under-sampling, MUSE exhibits severe stripe artifacts across the image due to signal loss from complex warping phases and fails to remove significant motion artifacts in the combined 3-direction, 3-average DWI (yellow arrow in **Fig. S15(b)**). While methods based on global phase smoothness priors (MUSSELS, S-LORAKS, PAIR) also show residual motion artifacts (yellow arrows in Fig. **S14(d)-(g)**), maybe because abdominal motion phases violate their inherent smoothness assumption. DONATE (1D low-rank in readout) and LLR (patch low-rank in image-domain) demonstrate slightly superior motion artifact suppression as they require no global phase smoothness. However, DONATE and LLR suffer from low signal-to-noise ratio (SNR). Critically, all compared methods exhibit severe background noise residuals in their reconstructions.

Compared with other methods, the baseline (LoSP) of the proposed method has the better noise suppression performance and can effectively suppress motion artifacts. However, in some region, some signal loss can be observed maybe due to the tight low-rank constraints (purple arrows in **Fig. S15(i)**). Prompt learning (LoSP-Prompt) further improves the reconstruction SNR and protects the small tissue structures in the image (purple arrows in **Fig. S15(j)**). Also, LoSP-Prompt has the best suppression of background noise.

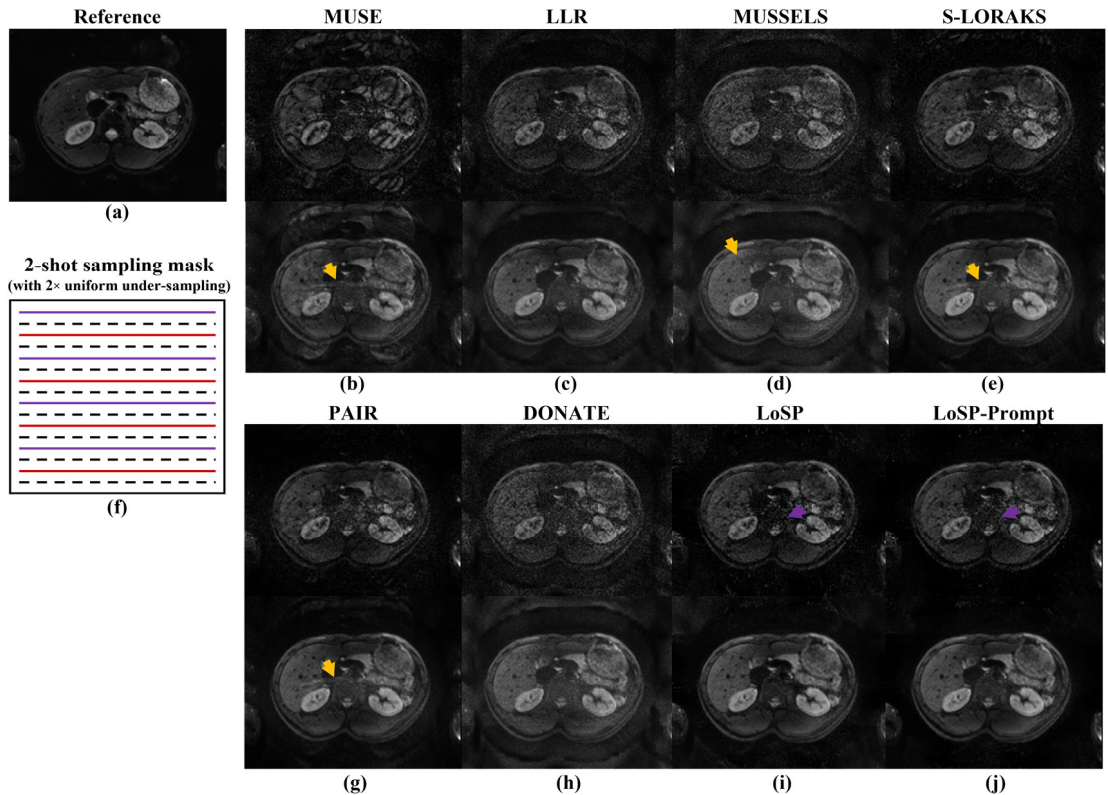

**Figure S15: Comparison study on the accelerated reconstructions (uniform under-sampling).** (a) is a reference image with a b value of 50 s/mm$^2$. (b)-(j) are the reconstructions of MUSE, LLR, MUSSELS, S-LORAKS, PAIR, DONATE, LoSP, and LoSP-Prompt, respectively. The first and third rows are the reconstructed DWI images of a single diffusion direction averaged once, and the second and fourth rows are DWI images combined by 3 directions and 3 averages. (f) is sampling mask. Note: The data is from **DATASET II**, acquired with Neusoft 3T, Universal scanner, matrix size 180×144, b-value 800 s/mm$^2$, subject ID is **HS#10**.

In partial Fourier under-sampling reconstructions, both DONATE and LLR exhibit significant image blurring and loss of fine structures, such as the obscured small liver blood vessels (purple arrows in **Fig. S16(a-b)**). This blurring arises because LLR, as an image-domain denoising method, inherently smooths details, an effect exacerbated by missing high-frequency k-space data. Similarly, DONATE, reliant on tight image-domain support without exploiting k-space conjugate symmetry, fails to recover high-frequency information, leading to further blur.

In contrast, the proposed LoSP and LoSP-Prompt methods demonstrate superior reconstruction fidelity, preserving small tissue structures (purple arrows in **Fig. S6(c-d)**) and effectively mitigating blur. Notably, the prompt-learning enhanced LoSP-Prompt further suppresses background noise.

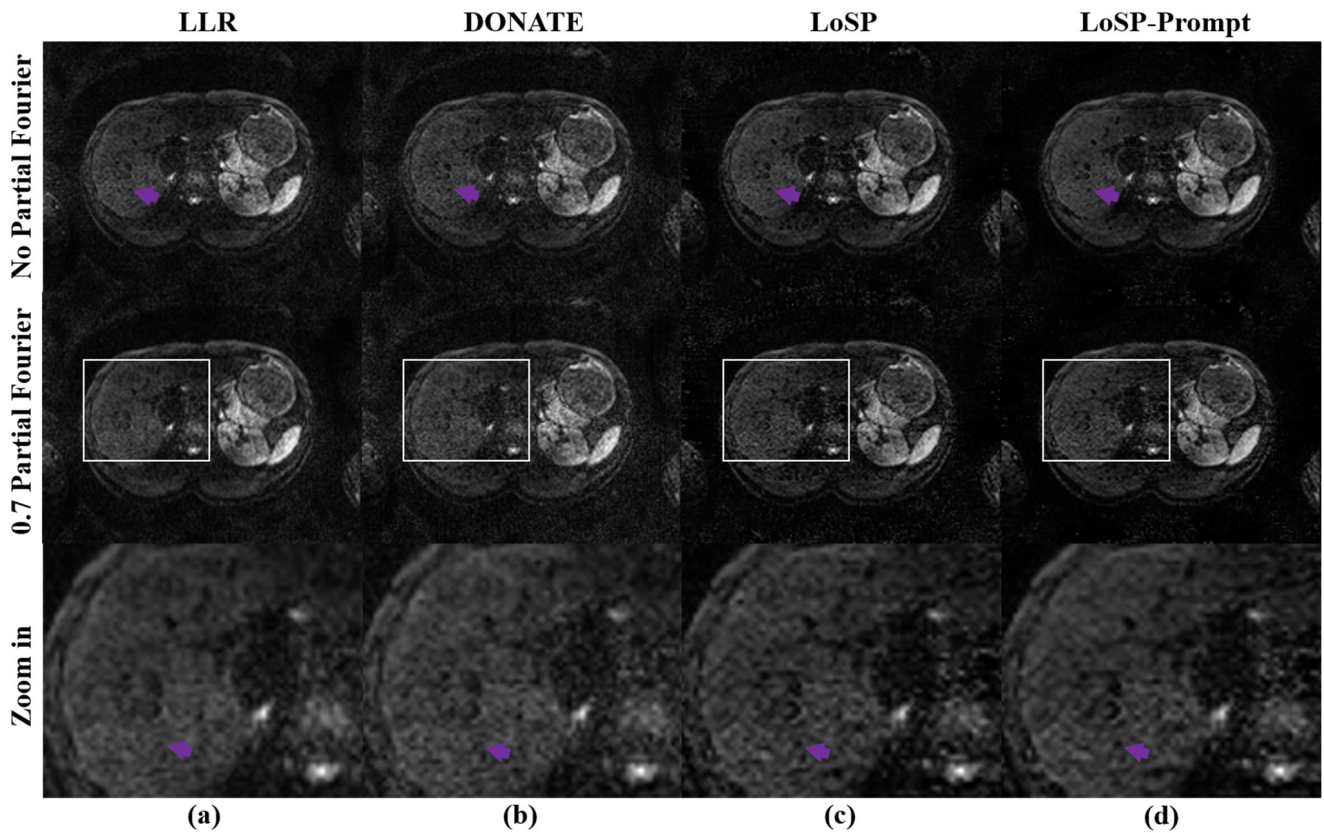

**Figure S16: Comparison study on the accelerated reconstructions (retrospectively partial Fourier under-sampling).** (a)-(d) are the reconstructed DWI images of a single diffusion direction averaged once by LLR, DONATE, LoSP, and LoSP-Prompt, respectively. The first row is the result without half-Fourier under-sampling, the second row is the result of half-Fourier under-sampling with a sampling rate of 0.7, and the third row is a local enlarged view of the second row. Note: The data is from **DATASET II**, acquired with Neusoft 3T, Universal scanner, matrix size 180×144, b value 800 s/mm$^2$, subject ID is **HS#10**.

## Note 7. Ablation study

Ablation experiments are conducted to compare the performance between the proposed bidirectional 1D low-rank algorithm LoSP and the existing 1D low-rank reconstruction method DONATE[7].

For fair comparison, two unidirectional LoSP models are included as:

$$(LoSP(PE)) \quad \min_{\mathbf{X}} \frac{\lambda}{2}\left\|\mathbf{Y}-\mathcal{UFC}\mathcal{F}^{-1}\mathbf{X}\right\|_F^2 + \sum_{m=1}^{N}\left\|\mathcal{H}\mathcal{S}_n^{PE}\left(\mathcal{F}_{RO}^{-1}\mathbf{X}\right)\right\|_*, \tag{S.22}$$

$$(LoSP(RO)) \quad \min_{\mathbf{X}} \frac{\lambda}{2}\left\|\mathbf{Y}-\mathcal{UFC}\mathcal{F}^{-1}\mathbf{X}\right\|_F^2 + \sum_{m=1}^{M}\left\|\mathcal{H}\mathcal{S}_m^{RO}\left(\mathcal{F}_{PE}^{-1}\mathbf{X}\right)\right\|_*, \tag{S.23}$$

$$(LoSP(RO \& PE)) \quad \min_{\mathbf{X}} \frac{\lambda}{2}\left\|\mathbf{Y}-\mathcal{UFC}\mathcal{F}^{-1}\mathbf{X}\right\|_F^2 + \sum_{m=1}^{M}\left\|\mathcal{H}\mathcal{S}_m^{RO}\left(\mathcal{F}_{PE}^{-1}\mathbf{X}\right)\right\|_* + \sum_{m=1}^{N}\left\|\mathcal{H}\mathcal{S}_n^{PE}\left(\mathcal{F}_{RO}^{-1}\mathbf{X}\right)\right\|_*, \tag{S.24}$$

The low-rank constraint is applied on frequency encoding dimension for LoSP (RO) and DONATE, in the phase encoding dimension for LoSP (PE), and both dimensions for LoSP (RO & PE). All models are solved using the ADMM algorithm.

LoSP (PE) fails to reconstruct a reasonable DWI image (**Fig. S17(b)**). This is because the multi-shot data is under-sampled in each frequency encoding line, and each phase encoding line is either fully sampled or not sampled. Thus, it is invalid to constrain only the phase encoding line.

Both LoSP (RO) and DONATE (RO) leads to reasonable DWI images in **Fig. S17(c)** and **Fig. S17(a)**, respectively. The former achieves better suppresses the noise and clearer image edges.

The proposed bidirectional LoSP (RO & PE) improves the signal-to-noise ratio (**Fig. S17(d)**) than LoSP (RO).

The proposed LoSP-Prompt (RO & PE) introduces prompt learning to automatically set the optimal singular value truncation threshold in the reconstruction, enabling much better signal-to-noise ratio and clearer edges **(Fig. S17(g))**.

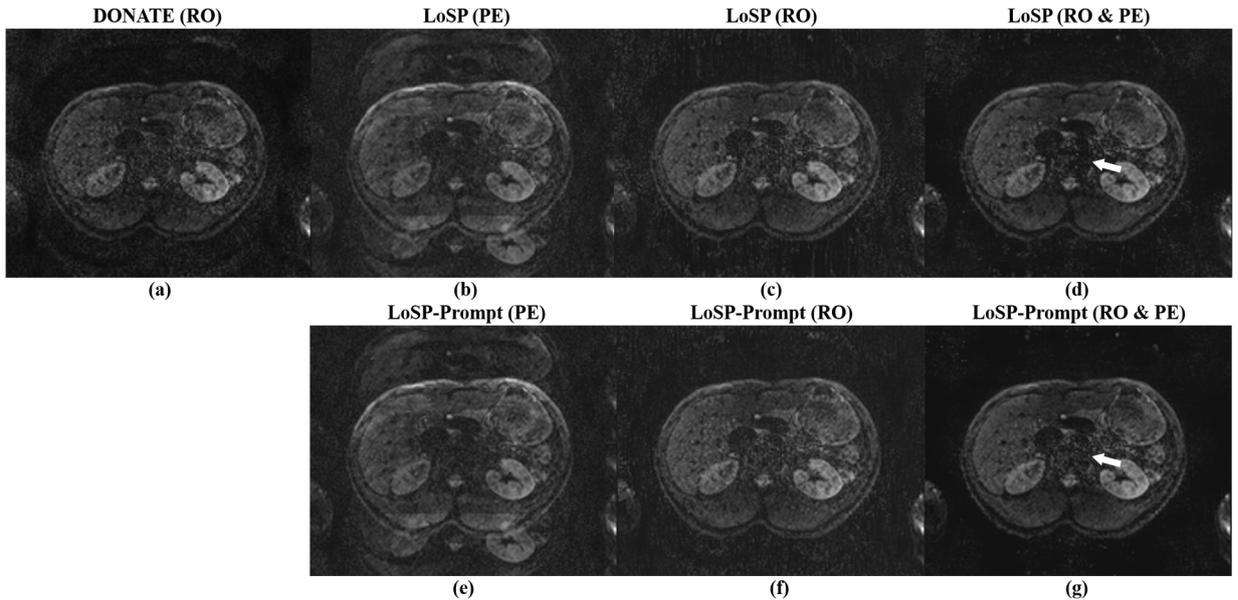

**Figure S17: Ablation study on 1D low-rankness.** (a)-(d) are reconstructed by DONATE (RO), LoSP (PE), LoSP (RO), and LoSP (RO & PE), respectively. (e)-(g) are reconstructed by LoSP-Prompt, corresponding to (b)-(d). Note: The data is from DATASET II, which is acquired with Neusoft 3T, Universal scanner, with matrix size 180×144 and b-value 800 s/mm$^2$, subject ID is **HS#10**.

## Note 8. Reader study details

This section gives the detailed scores of each radiologist.

### TABLE V. Reader scores (mean ± standard deviation) of DWI images

| Parts | Reader | Criterion | LLR | MUSSELS | LoSP-Prompt |
|---|---|---|---|---|---|
| Liver | #1 | Overall image quality | 2.90±1.12 | 2.36±1.48 | **3.46±0.91** |
| | | SNR | 2.82±1.14 | 2.28±1.50 | **3.39±0.95** |
| | | Artifact suppression | 2.93±1.12 | 2.39±1.47 | **3.48±0.92** |
| | #2 | Overall image quality | 3.52±0.18 | 3.53±0.12 | **4.45±0.15** |
| | | SNR | 3.67±0.32 | 3.74±0.23 | **4.52±0.18** |
| | | Artifact suppression | 3.42±0.21 | 3.48±0.14 | **4.43±0.13** |
| | #3 | Overall image quality | 3.47±0.24 | 3.57±0.23 | **3.94±0.18** |
| | | SNR | 3.56±0.31 | 3.57±0.33 | **3.94±0.19** |
| | | Artifact suppression | 3.17±0.33 | 3.48±0.25 | **3.98±0.10** |
| | Total | Overall image quality | 3.24±0.84 | 2.78±1.14 | **3.85±0.73** |
| | | SNR | 3.25±0.89 | 2.79±1.19 | **3.85±0.79** |
| | | Artifact suppression | 3.19±0.84 | 2.76±1.13 | **3.89±0.75** |
| Kidney | #1 | Overall image quality | 4.42±0.10 | 4.31±0.15 | **4.40±0.13** |
| | | SNR | 4.41±0.13 | 4.34±0.12 | **4.37±0.14** |
| | | Artifact suppression | 4.40±0.10 | 4.23±0.21 | **4.39±0.14** |
| | #2 | Overall image quality | 4.17±0.90 | 4.01±0.84 | **4.48±1.02** |
| | | SNR | 4.38±0.46 | 4.21±0.41 | **4.81±0.46** |
| | | Artifact suppression | 4.40±0.58 | 4.17±0.54 | **4.57±0.57** |
| | #3 | Overall image quality | 3.27±0.71 | 3.10±0.76 | **3.58±0.68** |
| | | SNR | 3.25±0.69 | 3.09±0.73 | **3.57±0.69** |
| | | Artifact suppression | 3.29±0.74 | 3.11±0.79 | **3.61±0.68** |
| | Total | Overall image quality | 3.95±0.82 | 3.80±0.83 | **4.15±0.81** |
| | | SNR | 4.01±0.72 | 3.88±0.74 | **4.25±0.70** |
| | | Artifact suppression | 4.03±0.76 | 3.84±0.76 | **4.19±0.66** |

| Parts | Reader | Criterion | LLR | MUSSELS | LoSP-Prompt |
|---|---|---|---|---|---|
| Uterus | #4 | Overall image quality | 2.43±0.64 | 2.20±0.62 | **3.01±0.56** |
| | | SNR | 2.43±0.76 | 2.12±0.65 | **2.95±0.68** |
| | | Artifact suppression | 2.16±0.78 | 2.02±0.77 | **2.92±0.81** |
| | #5 | Overall image quality | 2.48±0.66 | 2.29±0.68 | **2.85±0.46** |
| | | SNR | 2.54±0.97 | 2.05±0.75 | **2.70±0.69** |
| | | Artifact suppression | 1.95±0.77 | 1.89±0.92 | **2.73±0.73** |
| | #6 | Overall image quality | 2.56±0.77 | 2.18±0.69 | **3.13±0.82** |
| | | SNR | 2.58±0.75 | 2.19±0.72 | **3.12±0.90** |
| | | Artifact suppression | 2.44±0.92 | 2.16±0.69 | **3.10±1.04** |
| | Total | Overall image quality | 2.43±0.64 | 2.20±0.62 | **3.01±0.56** |
| | | SNR | 2.43±0.76 | 2.12±0.65 | **2.95±0.68** |
| | | Artifact suppression | 2.16±0.78 | 2.02±0.77 | **2.92±0.81** |
| Sacroiliac | #7 | Overall image quality | 3.34±0.26 | 3.32±0.49 | **3.87±0.21** |
| | | SNR | 3.30±0.22 | 3.32±0.49 | **3.90±0.20** |
| | | Artifact suppression | 3.31±0.22 | 3.32±0.49 | **3.90±0.23** |
| | #8 | Overall image quality | 1.15±0.54 | 1.48±1.09 | **3.92±0.41** |
| | | SNR | 1.13±0.51 | 1.46±1.10 | **3.78±0.84** |
| | | Artifact suppression | 1.15±0.51 | 1.51±1.12 | **3.90±0.45** |
| | #9 | Overall image quality | 1.49±0.79 | 1.95±1.64 | **4.26±0.44** |
| | | SNR | 1.25±0.66 | 1.74±1.48 | **3.76±0.81** |
| | | Artifact suppression | 0.87±0.44 | 1.52±1.34 | **3.48±0.53** |
| | Total | Overall image quality | 1.99±1.12 | 2.25±1.40 | **4.02±0.40** |
| | | SNR | 1.90±1.12 | 2.17±1.37 | **3.81±0.68** |
| | | Artifact suppression | 1.78±1.17 | 2.12±1.34 | **3.76±0.46** |

| Parts | Reader | Criterion | LLR | MUSSELS | LoSP-Prompt |
|---|---|---|---|---|---|
| Spinal cord | #7 | Overall image quality | 3.57±0.27 | 3.49±0.31 | **3.76±0.34** |
| | | SNR | 3.54±0.28 | 3.51±0.33 | **3.74±0.35** |
| | | Artifact suppression | 3.54±0.29 | 3.51±0.34 | **3.78±0.36** |
| | #8 | Overall image quality | 3.56±0.38 | 3.54±0.17 | **4.03±0.25** |
| | | SNR | 3.57±0.39 | 3.53±0.17 | **4.02±0.26** |
| | | Artifact suppression | 3.59±0.39 | 3.50±0.17 | **4.01±0.25** |
| | #9 | Overall image quality | 3.85±0.67 | 3.47±0.96 | **4.02±0.48** |
| | | SNR | 3.57±0.31 | 3.32±0.62 | **3.69±0.37** |
| | | Artifact suppression | 2.69±0.28 | 2.24±0.70 | **2.72±0.41** |
| | Total | Overall image quality | 3.66±0.48 | 3.50±0.58 | **3.94±0.38** |
| | | SNR | 3.56±0.32 | 3.45±0.42 | **3.82±0.36** |
| | | Artifact suppression | 3.27±0.52 | 3.08±0.76 | **3.50±0.66** |
| Knee | #7 | Overall image quality | 3.71±0.32 | 3.46±0.24 | **3.67±0.21** |
| | | SNR | 3.70±0.32 | 3.49±0.23 | **3.69±0.21** |
| | | Artifact suppression | 3.71±0.32 | 3.37±0.68 | **3.68±0.26** |
| | #8 | Overall image quality | 2.88±0.64 | 2.49±0.68 | **3.40±0.51** |
| | | SNR | 2.86±0.64 | 2.47±0.69 | **3.37±0.53** |
| | | Artifact suppression | 2.85±0.62 | 2.39±0.64 | **3.40±0.51** |
| | #9 | Overall image quality | **3.61±0.65** | 2.51±0.49 | 3.57±0.61 |
| | | SNR | **3.18±0.81** | 2.29±0.74 | 2.94±0.81 |
| | | Artifact suppression | 2.81±0.57 | 1.80±0.71 | **3.16±0.56** |
| | Total | Overall image quality | 3.40±0.67 | 2.82±0.68 | **3.55±0.48** |
| | | SNR | 3.25±0.71 | 2.75±0.79 | **3.34±0.64** |
| | | Artifact suppression | 3.12±0.66 | 2.52±0.94 | **3.41±0.51** |

| Parts | Reader | Criterion | LLR | MUSSELS | LoSP-Prompt |
|---|---|---|---|---|---|
| Brain tumor | #9 | Overall image quality | 2.81±0.52 | **4.07±0.75** | 3.94±0.55 |
| | | SNR | 3.06±0.65 | **4.06±0.73** | 4.03±0.47 |
| | | Artifact suppression | 2.59±0.47 | 3.83±1.11 | **3.87±0.59** |
| | #10 | Overall image quality | 3.56±0.36 | 3.86±0.44 | **3.89±0.34** |
| | | SNR | 3.59±0.35 | 3.84±0.44 | **3.84±0.32** |
| | | Artifact suppression | 3.40±0.55 | 3.95±0.53 | **4.06±0.36** |
| | #11 | Overall image quality | 3.05±1.12 | **3.65±1.27** | 3.40±1.40 |
| | | SNR | 3.34±0.92 | 3.71±1.00 | **3.73±0.51** |
| | | Artifact suppression | 2.80±0.91 | 3.76±1.24 | **4.30±0.63** |
| | Total | Overall image quality | 3.14±0.80 | **3.86±0.90** | 3.74±0.92 |
| | | SNR | 3.33±0.71 | 3.87±0.77 | **3.87±0.46** |
| | | Artifact suppression | 2.93±0.75 | 3.85±1.00 | **4.08±0.56** |

Note: Best performance is marked with bold letters. The reader study is performed through our cloud computing evaluation platform, CloudBrain-ReconAI[10], which is free to access at https://csrc.xmu.edu.cn/CloudBrain.html.

## Note 9. Comparison of reconstruction parameters

In all experiments, as shown in **Table VI**, all methods are given with a set of optimized subject-specific parameters (not image-specific) to best balance motion artifacts removal and noise suppression. Among these methods, codes of IRIS[11], MUSE[12] and S-LORAKS[4] are reproduced according to the corresponding papers; MUSSELS, DONATE, LLR, and PAIR are provided by original authors.

**TABLE VI. Reconstruction parameters**

| Methods | Main parameters | Number (settings) |
|---|---|---|
| MUSE | Iteration 1: Number of iterations in step 1<br>Iteration 2: Number of iterations in step 2 | 2 (Manual) |
| LLR (POCS) | Iteration: Number of iterations<br>Lambda: Regularization parameter for locally low-rank term | 2 (Manual) |
| S-LORAKS (POCS) | Iteration: Number of iterations<br>Radius: Radius for constructing block Hankel<br>Saved rank: Number of saved ranks in singular value decomposition<br>Lambda: Regularization parameter for nuclear norm | 4 (Manual) |
| MUSSELS (IRLS) | Iteration 1: Number of outer iterations<br>Iteration 2: Number of inner iterations<br>Ksize: Window size for constructing block Hankel<br>Lambda: Regularization parameter for Frobenius norm | 4 (Manual) |
| PAIR (POCS) | Iteration: Number of iterations<br>Radius: Radius for constructing block Hankel<br>Saved rank: Number of saved ranks in singular value decomposition<br>Lambda 1: Regularization parameter for nuclear norm<br>Lambda 2: Regularization parameter for weighted total variation norm | 4 (Manual) |
| DONATE (POCS) | Iteration: Number of iterations<br>Lsize: Length for constructing Hankel<br>Saved rank: Number of saved ranks in singular value decomposition<br>Lambda: Regularization parameter for nuclear norm | 3 (Manual)<br>+ $M$ (Manual) |
| LoSP (ADMM) | Iteration: Number of iterations, default to 20.<br>Lsize: Length for constructing Hankel, default to 10.<br>Lambda: Regularization parameter for nuclear norm, default to 1.<br>Saved ranks: Numbers of saved ranks in singular value decomposition | 3 (Default)<br>+ $N$ (Manual)<br>+ $M$ (Manual) |
| LoSP-Prompt (ADMM) | Iteration: Number of iterations, default to 20.<br>Lsize: Length for constructing Hankel, default to 10.<br>Lambda: Regularization parameter for nuclear norm, default to 1.<br>Saved ranks: Numbers of saved ranks in singular value decomposition | 3 (Default)<br>+ $N$ (Automatically)<br>+ $M$ (Automatically) |

Note: POCS, IRLS, and ADMM are short for Projection onto Convex Sets, Iterative Reweighted Least Squares, and Alternating Direction Method of Multipliers, respectively. $N$ and $M$ are the numbers of phase encoding and readout lines.

## Note 10. Recovery of 1D signals with different low-rankness

In this section, we test the performance of LoSP-Prompt on separately 1D signal reconstructions to evaluate its ability to handle 1D signals with different 1D low-rankness (**Fig. S18**). Two 1D signals, $S_1$ and $S_2$, are selected from regions with 1-order and 5-order phase (**Fig. S18(a)**), respectively. The 1D signal $S_1$ has a significantly better low-rankness than $S_2$ (**Fig. S18(b)**).

The results show that, LoSP-Prompt can well reconstruct the 1D signal with good low-rank performance (**Fig. S18(c, d)**). Moreover, it can also recover the signal with poor low-rank performance reliably (**Fig. S18(e, f)**) with a lower PSNR.

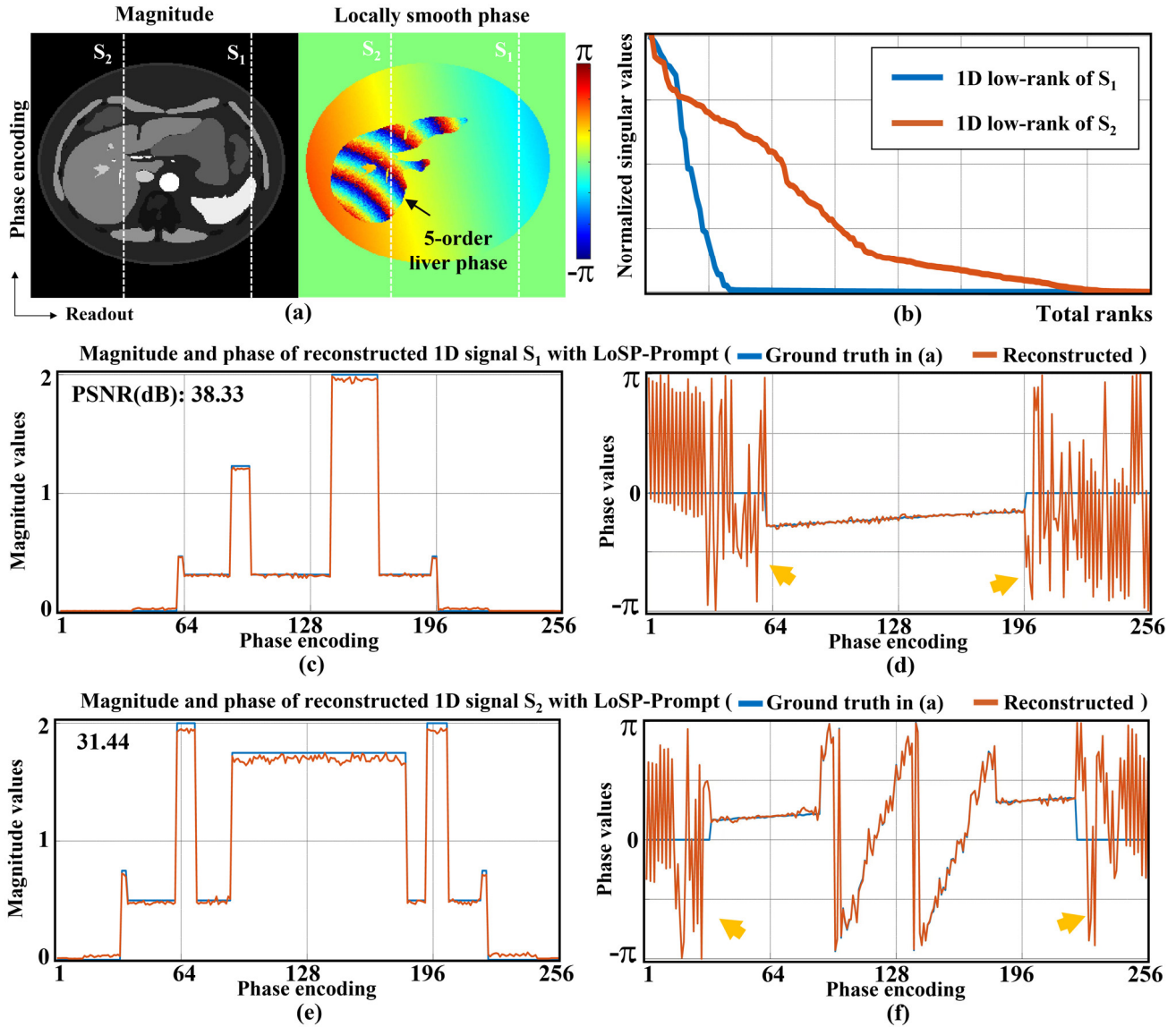

**Figure S18: Reconstruction of 1D signals ($S_1$ and $S_2$) with distinct low-rankness. (a)** are the magnitude and phase of synthesized 4-shot DWI. **(b)** is the 1D low-rank of 1D signals with 1-order ($S_1$) and 5-order ($S_2$) phase, respectively. **(c)** and **(d)** are the magnitude and phase of reconstructed 1D signal $S_1$ with LoSP-Prompt, respectively. **(e)** and **(f)** are the magnitude and phase of reconstructed 1D signal $S_2$ with LoSP-Prompt, respectively. Note: The yellow arrows indicate normal oscillations of reconstructed phase in the background area (Magnitude = 0, phase = 0); The PSNRs are marker at the top of **(c)** and **(e)**.

## Note 11. Pathology reports of the patient (PS#1) with hepatocellular carcinoma

We present the pathology of the patient (PS#1) with hepatocellular carcinoma (Fig. **S19**) as follows:

**Gross Description:** Liver mass. Multiple fragments of grayish-white and grayish-yellow tissue, aggregating to 0.5 × 0.3 × 0.1 cm$^3$ in total dimension. Soft in consistency.

**Pathological Diagnosis:** (Liver mass, needle biopsy) Well-differentiated hepatocellular neoplasm. Increased cellular density with mild atypia. Combined with immunohistochemical findings, the features favor well-differentiated hepatocellular carcinoma.

**Immunohistochemistry:** AFP (-), CD34 (Focal sinusoidal capillarization), CEA (Focal +), CK18 (+), CK19 (-), CK7 (-), CK8 (+), GPC3 (-), HepPar-1 (+), Ki67 (Scattered cells +), MUC-1 (-), GS (+), HSP70 (+), Arginase-1 (Focal +).

**Special Stain:** Reticulin stain, showing thickened hepatic plates/multilayering of hepatocytes.

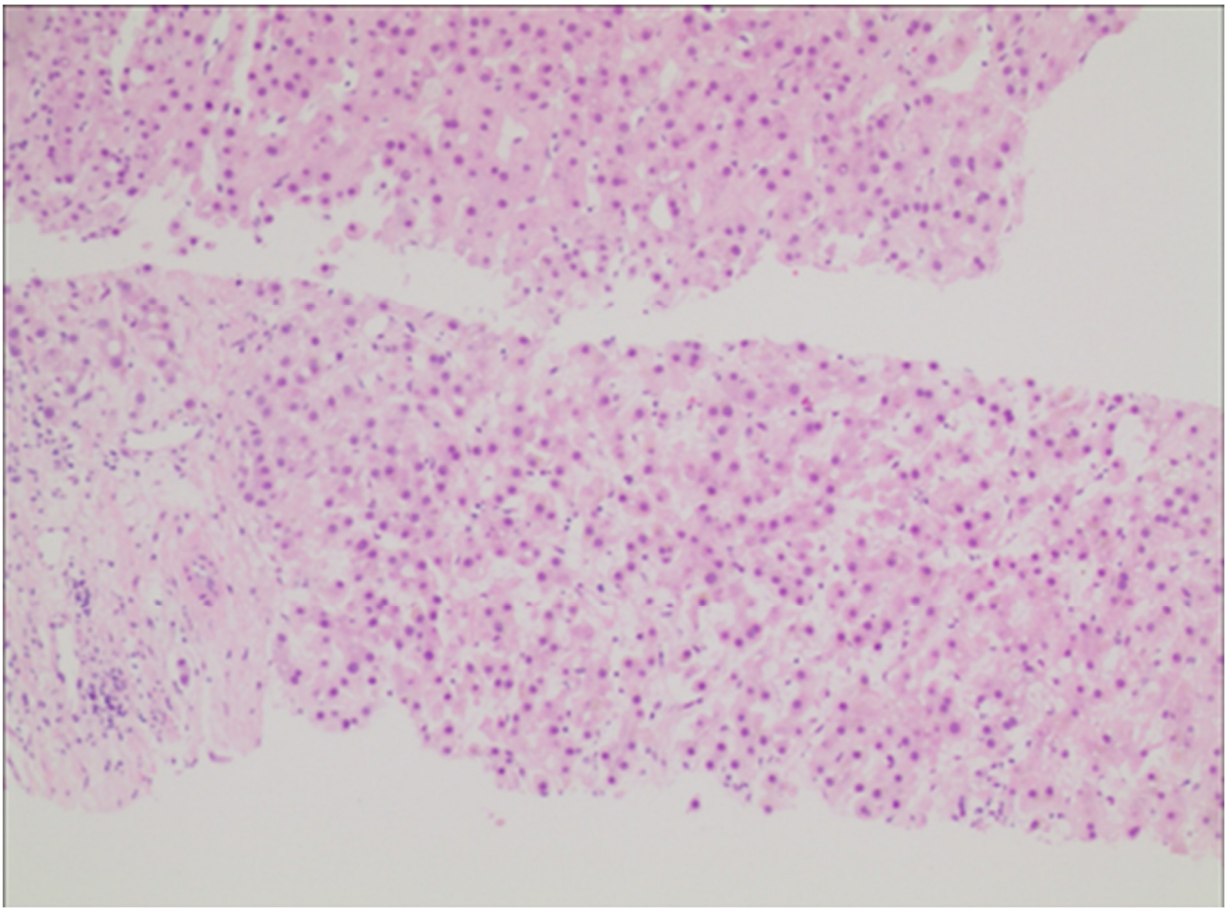

**Figure S19: Pathological section staining image of liver tissue from patient subject PS#1.**